%% file: main_sop_bench_arxiv_2026-feb.tex
\documentclass[sigconf,authorversion]{acmart}

\AtBeginDocument{%
  }

\setcopyright{none}

\renewcommand\footnotetextcopyrightpermission[1]{}
\usepackage{afterpage}
\usepackage{booktabs}
\usepackage{colortbl}
\usepackage{dblfloatfix}
\usepackage{listings}
\usepackage{pifont}
\usepackage{tcolorbox}
\usepackage{xcolor}
\usepackage{capt-of}

\lstdefinelanguage{json}{
  basicstyle=\ttfamily\small,
  numbers=left,
  numberstyle=\tiny\color{gray},
  stepnumber=1,
  numbersep=8pt,
  showstringspaces=false,
  breaklines=true,
  frame=single,
  backgroundcolor=\color{gray!5}
}

\lstdefinelanguage{yaml}{
  basicstyle=\ttfamily\small,
  numbers=left,
  numberstyle=\tiny\color{gray},
  stepnumber=1,
  numbersep=8pt,
  showstringspaces=false,
  breaklines=true,
  frame=single,
  backgroundcolor=\color{gray!5}
}

\definecolor{goodgreen}{RGB}{0,150,0}
\definecolor{badred}{RGB}{200,0,0}
\definecolor{soprow}{RGB}{235,255,235}
\definecolor{headergray}{gray}{0.92}

\newcommand{\cmark}{\textcolor{goodgreen}{\ding{51}}}
\newcommand{\xmark}{\textcolor{badred}{\ding{55}}}

\newcommand{\amazonaffil}{\affiliation{
  \institution{Amazon}
  \country{USA}
}}

\begin{document}

\settopmatter{authorsperrow=4}

\title[SOP-Bench]{SOP-Bench: Complex Industrial SOPs for Evaluating LLM Agents}


\author{Subhrangshu Nandi}
\authornote{Primary Authors}
\email{subhrn@amazon.com}
\amazonaffil

\author{Arghya Datta}
\authornotemark[1]
\email{arghya.dat@gmail.com}
\affiliation{
  \institution{ex-Amazon}
  \country{USA}
}

\author{Rohith Nama}
\authornotemark[1]
\email{rohinama@amazon.com}
\amazonaffil

\author{Udita Patel}
\authornotemark[1]
\email{patudita@amazon.com}
\amazonaffil

\author{Nikhil Vichare}
\authornotemark[1]
\email{nvichare@amazon.com}
\amazonaffil

\author{Indranil Bhattacharya}
\email{bindrani@amazon.com}
\authornotemark[1]
\amazonaffil

\author{Prince Grover}
\email{pringrov@amazon.com}
\amazonaffil

\author{Shivam Asija}
\email{asijashi@amazon.com}
\amazonaffil

\author{Giuseppe Carenini}
\email{carenini@amazon.com}
\amazonaffil

\author{Wei Zhang}
\email{wzhngm@amazon.com}
\amazonaffil

\author{Arushi Gupta}
\email{arushimg@amazon.com}
\amazonaffil

\author{Sreyoshi Bhaduri}
\email{drsre@amazon.com}
\amazonaffil

\author{Jing Xu}
\email{xuzjx@amazon.com}
\amazonaffil

\author{Shayan Ray}
\email{shayanray2018@gmail.com}
\affiliation{
  \institution{ex-Amazon}
  \country{USA}
}

\author{Huzefa Raja}
\email{huzeraja@amazon.com}
\amazonaffil

\author{Aaron Chan}
\email{chaaron@amazon.com}
\amazonaffil

\author{Esther Xu Fei}
\email{exf@amazon.com}
\amazonaffil

\author{Gaoyuan Du}
\email{gdu@amazon.com}
\amazonaffil

\author{Zuhaib Akhtar}
\email{zuhaibak@amazon.com}
\amazonaffil

\author{Harshita Asnani}
\email{asnaharn@amazon.com}
\amazonaffil

\author{Weian Chen}
\email{weiac@amazon.com}
\amazonaffil

\author{Ming Xiong}
\email{xiongmin@amazon.com}
\amazonaffil

\author{Francesco Carbone}
\email{carbonef@amazon.com}
\amazonaffil

\author{Jeetu Mirchandani}
\email{jeetu@amazon.com}
\amazonaffil

\renewcommand{\shortauthors}{Nandi et al.}


\input{kdd_tex_files/sec00_abstract}

\maketitle

\fancyhead{}
\fancyfoot{}
\renewcommand{\headrulewidth}{0pt}
\renewcommand{\footrulewidth}{0pt}


\clearpage

\twocolumn[{
    \centering
    \includegraphics[width=\textwidth]{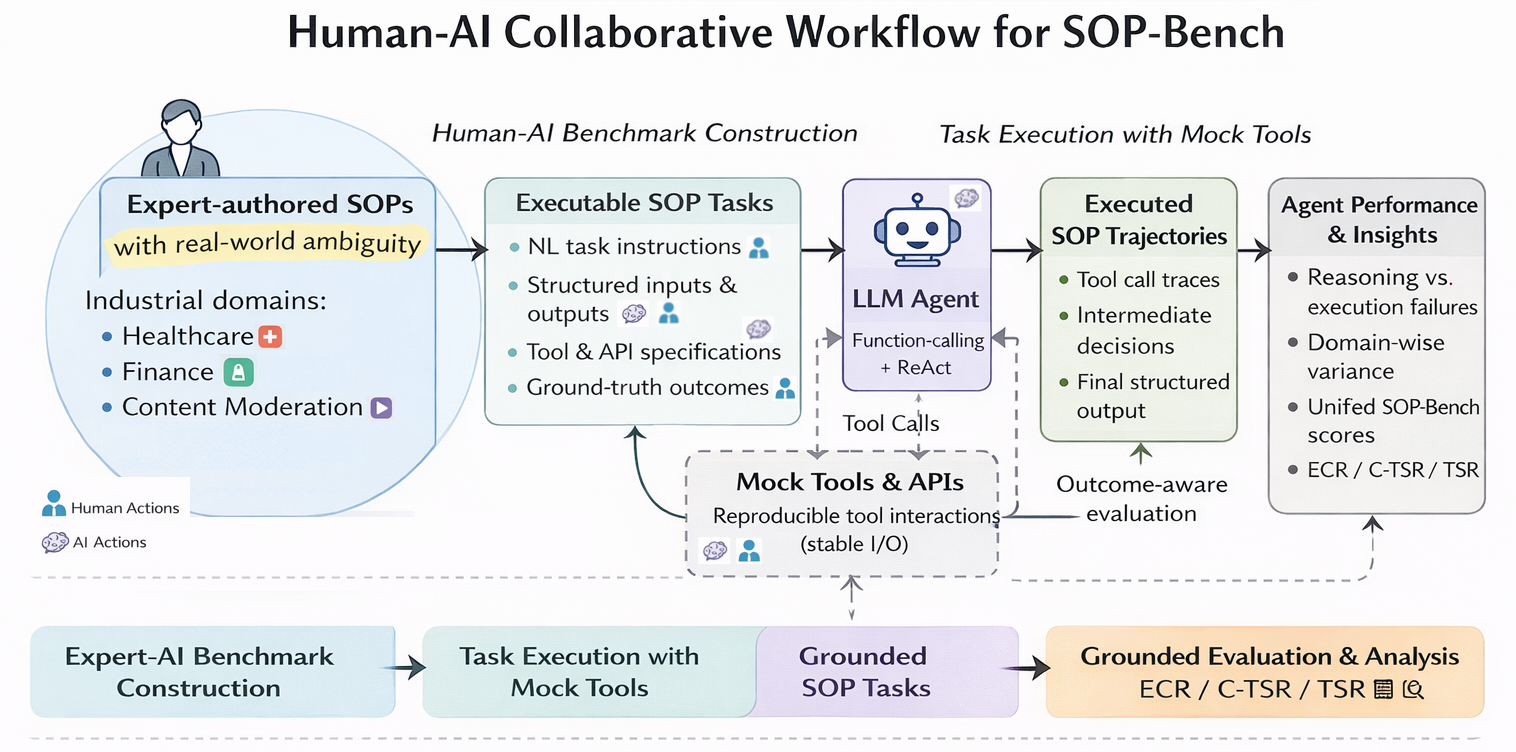}
    \captionof{figure}{SOP-Bench evaluation overview. Realistic business process SOPs authored by human experts across diverse domains are converted into executable task instances with structured tool/API interfaces and ground-truth outputs. LLM agents execute tasks via reproducible tool interactions, producing execution trajectories evaluated using grounded, outcome-aware metrics (ECR, C-TSR, TSR).}
    \label{fig:sop_bench_teaser}
    \vspace{0.3cm}
}]


\input{kdd_tex_files/sec01_intro}
\input{kdd_tex_files/sec02_related_work}

\input{kdd_tex_files/sec03_benchmark_design}
\input{kdd_tex_files/sec04_agents_for_automating_sops}
\input{kdd_tex_files/sec05_results}
\input{kdd_tex_files/sec06_conclusion}


\balance
\bibliographystyle{plainnat}
\bibliography{references}


\appendix
\section*{Appendix}

\input{kdd_tex_files/appendix_gen-ai_use}
\input{kdd_tex_files/appendix_agents}
\input{kdd_tex_files/appendix_sop_details}

\input{kdd_tex_files/appendix_figures}
\input{kdd_tex_files/appendix_prompt}
\input{kdd_tex_files/appendix_sop_example}

\end{document}

%% file: kdd_tex_files/sec00_abstract.tex
\begin{abstract}
LLM-based agents struggle to execute complex, multi-step Standard Operating Procedures (SOPs) that are fundamental to industrial automation. Existing benchmarks 
fail to capture the procedural complexity and tool orchestration demands of real-world workflows. We introduce \textbf{SOP-Bench}, a benchmark of 2,000+ tasks 
from human expert-authored SOPs across 12 business domains (healthcare, logistics, finance, content moderation, etc.). Using a human-AI collaborative framework, 
experts crafted authentic SOPs while AI generated artifacts (tools, APIs, datasets), all human-validated, yielding realistic tasks with executable interfaces and 
ground-truth outputs.

SOP-Bench serves as a research enabler for systematically investigating agent architectures, model capabilities, and deployment considerations across diverse 
procedural tasks. We demonstrate its utility through illustrative experiments with a subset of frontier models across Function-Calling (FC) and ReAct agents, revealing 
critical insights. For example, (1) newer models do not guarantee better performance - Claude 4 family outperforms Claude 4.5 family on ReAct tasks (Claude 4 Opus: 72.4\% 
vs. Claude 4.5 Sonnet: 63.3\% task success rate), demonstrating that production upgrades require validation; (2) no single model-agent combination dominates: best 
performances range from 57\% to 100\% depending on domain. These examples illustrate how SOP-Bench enables isolating and studying specific dimensions of agent performance without costly production experiments. Our goal is not 
to rank model capabilities or build optimal agents, but to provide a rigorous evaluation framework that enables the researchers and practitioners to systematically investigate agent 
design choices, model selection, and deployment strategies. We release the benchmark at {\url{https://github.com/amazon-science/sop-bench}}.
\end{abstract}

%% file: kdd_tex_files/sec01_intro.tex

\section{Introduction}
Standard Operating Procedures (SOPs) are central to reliable operations across industries, encoding domain expertise, compliance rules, and decision logic into structured workflows \cite{de2009standard, dumas2018fundamentals}. Recent advances in large language models (LLMs) have sparked growing interest in automating SOP execution using LLM-based agents \cite{wang2023autogpt, yao2023react, chase2022langchain, gschwind2025}. However, deploying such agents in production exposes challenges that are not captured by existing benchmarks such as Gorilla~\cite{patil2023gorilla}, API-Bank~\cite{li-etal-2023-api}, or ComplexBench~\cite{wen2024complexbench}.

\begin{table}[!hb]
    \vspace{-0.1cm}
    \caption{Diversity of SOPs in industry, assessed by GPT-4-turbo \cite{openai2024chatgpt} and validated by human associates with extensive cross-domain SOP review experience.}
    \centering
    \small
    \definecolor{lightgray}{gray}{0.9}
    \begin{tabular}{l|c|c}
        \hline
        \textbf{Criteria} & \textbf{SLUHN} & \textbf{UNTHS} \\
        \hline
        Length & Long (7 pgs) & Short (3 pgs) \\
        Ambiguity & Low & Moderate \\
        Step complexity & High & Low-Mod \\
        Implicit knowledge & High & Low-Mod \\
        Modularity & High & Moderate \\
        \hline
    \end{tabular}
    \label{tab:sop_comparison}
    \vspace{-0.1cm}
\end{table}

These challenges are particularly acute in knowledge operations like verification services, support ticket triaging, and customer support, where SOPs 
involve ambiguous language, implicit domain knowledge, and complex decision trees \cite{roberts2019behind}. For example, consider the two patient 
registration SOPs in appendix \ref{appendix:real_sop}, one from St. Luke’s University Health Network 
(SLUHN)\footnote{\url{https://www.slhn.org/-/media/slhn/Research/File/PDF/SOPs-and-Policies/SOP-302Patient-Registration-and-Ongoing-Subject-MgmtV30CLEAN71216.ashx}}, and one from University of North Texas Health Science Center (UNTHS) \footnote{\url{https://www.unthsc.edu/wp-content/uploads/sites/23/Patient_Registration.pdf}}. 
Table \ref{tab:sop_comparison} shows how different two SOPs of the same domain can be. 

\begin{figure}[!ht]
    \vspace{-0.5cm}
    \centering
    \includegraphics[width=\columnwidth]{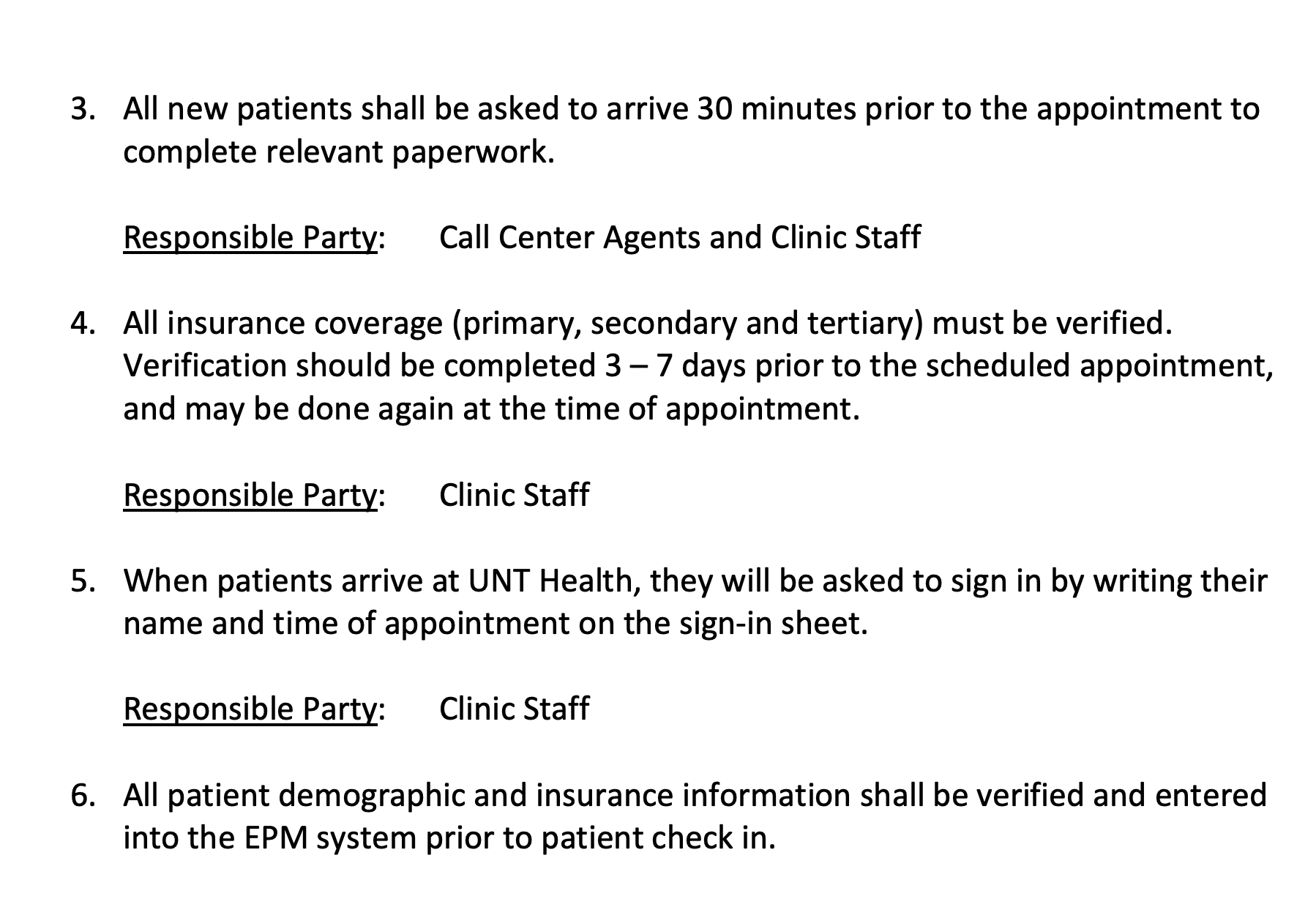}
    \caption{Example of instructions that can be simple for humans but confusing for LLMs. In this excerpt of page 2 of the UNHTS-SOP (appendix \ref{appendix:real_sop}), we can see that steps 4 and 6 both have instructions on verifying the insurance, but does not have explicit information on how to verify it. Moreover, it is not clear why there are two steps of insurance verifications. For human associates with implicit knowledge of the domains this might not be as ambiguous, while our experiments indicate that it is  for LLMs. In our experience, this is a common occurrence in business process SOPs.}
    \Description{Screenshot of a section of the UNHTS SOP showing numbered steps. Two different steps mention verifying insurance, but neither specifies how to verify or why verification is repeated.}
    \vspace{-0.5cm}
    \label{fig:example_sop}
\end{figure}

Such business knowledge operations differ fundamentally from traditional process automation tasks \cite{ribeiro2020auditing}. They require contextual interpretation, multi-step information synthesis, and judgment under ambiguity, often with evolving guidelines and subjective classifications \cite{roberts2019behind, gordon2022jury}. Figure~\ref{fig:example_sop} shows a representative example: instructions that are straightforward for human operators with implicit domain knowledge but ambiguous for LLM agents. Existing evaluation frameworks typically isolate individual capabilities—such as tool use \cite{huang2024metatool}, planning \cite{yao2023react}, or instruction following \cite{wang2022supernaturalinstructions}—using clean, synthetic prompts that fail to reflect the messiness of real-world SOPs.

While recent advances in LLMs have enabled sophisticated instruction following and tool use \cite{wang2022supernaturalinstructions, yao2023react}, 
automating SOP execution poses unique challenges that existing benchmarks fail to capture. Current evaluation frameworks primarily focus on isolated 
capabilities—tool use \cite{huang2024metatool}, planning \cite{yao2023react}, or instruction following \cite{wang2022supernaturalinstructions}, 
using clean, synthetic prompts that sidestep the messiness of real-world procedures.


To address this gap, we introduce \textbf{SOP-Bench}, a comprehensive benchmark for evaluating LLM agents on realistic business SOP execution. SOP-Bench is constructed from human expert-authored SOPs spanning twelve diverse domains, including healthcare intake, hazardous goods classification, customer service, content moderation, and financial compliance. Industry experts authored authentic procedures reflecting real-world ambiguity, branching logic, and implicit knowledge requirements. To enable reproducible evaluation, we adopt a human–AI collaborative workflow in which AI models generate supporting artifacts (mock tools, APIs, and test datasets) that are subsequently validated by human experts. This process yields over 2,000 executable tasks with structured tool interfaces and grounded outputs that closely mirror production environments.

We make three primary contributions:
\textbf{(1) Human--AI collaborative benchmark construction framework:} We introduce a scalable methodology where industry experts author authentic SOPs while AI assists in generating executable artifacts, all human-validated for correctness and realism.
\textbf{(2) A comprehensive benchmark for systematic agent evaluation:} SOP-Bench provides 2,000+ tasks across 12 domains with varying procedural complexity, document length, branching logic, and tool usage, enabling controlled study of agent architectures and foundation models without costly production experimentation.
\textbf{(3) Empirical insights via illustrative experiments:} Through experiments and ablations on representative frontier models, we demonstrate how SOP-Bench reveals non-obvious trade-offs in architecture choice, reasoning cost, and domain generalization.

Our goal is not merely to rank models, but to establish SOP-Bench as an evaluation substrate for studying agent design, model selection, and deployment strategies under realistic procedural constraints. We release the full benchmark, baseline agents, and evaluation framework at \url{https://github.com/amazon-science/sop-bench}, enabling the community to extend SOP-Bench with new domains and contribute back to a shared evaluation ecosystem.

%% file: kdd_tex_files/sec02_related_work.tex
\begin{table}[!ht]
\vspace{-0.2cm}
\caption{SOP-Bench uniquely combines all seven capabilities essential for real-world industrial workflow automation, addressing critical gaps in existing agent benchmarks.}
\small
\centering
\setlength{\tabcolsep}{3.5pt}
\renewcommand{\arraystretch}{1.1}

\begin{tabular}{@{}p{2.15cm}ccccccc@{}}
\toprule
\rowcolor{headergray}
\textbf{Benchmark} &
\textbf{IF} &
\textbf{MS} &
\textbf{Tool} &
\textbf{Amb.} &
\textbf{Err.} &
\textbf{Dep.} &
\textbf{Ind.} \\
\midrule
AlpacaEval   & \cmark & \xmark & \xmark & \xmark & \xmark & \xmark & \xmark \\
FollowEval   & \cmark & \xmark & \xmark & \xmark & \xmark & \xmark & \xmark \\
ComplexBench & \cmark & \cmark & \xmark & \xmark & \xmark & \cmark & \xmark \\
InFoBench    & \cmark & \cmark & \xmark & \xmark & \xmark & \xmark & \xmark \\
Gorilla      & \cmark & \xmark & \cmark & \xmark & \xmark & \xmark & \xmark \\
API-Bank     & \cmark & \xmark & \cmark & \xmark & \xmark & \xmark & \xmark \\
AgentBench   & \cmark & \cmark & \cmark & \xmark & \xmark & \xmark & \xmark \\
PlanBench    & \cmark & \cmark & \cmark & \xmark & \xmark & \cmark & \xmark \\
ALFWorld     & \cmark & \cmark & \cmark & \xmark & \xmark & \cmark & \xmark \\
SOP-Maze     & \cmark & \cmark & \xmark & \cmark & \xmark & \cmark & \cmark \\
\midrule
\rowcolor{soprow}
\textbf{SOP-Bench} & \cmark & \cmark & \cmark & \cmark & \cmark & \cmark & \cmark \\
\bottomrule
\end{tabular}

\vspace{0.4ex}
\raggedright
\footnotesize
\textbf{Legend:} IF = Instruction Following; MS = Multi-step Tasks; Tool = Tool/API Integration; Amb. = Real-world Ambiguity; Err. = Error Handling; Dep. = Workflow Dependencies; Ind. = Industrial Context.
\vspace{-0.5cm}
\label{tab:benchmark_comparison}
\end{table}

\section{Related Work}

\textbf{LLM Agents for Planning and Tool Use.}
LLMs have enabled the emergence of AI agents capable of task planning, reasoning, and tool use. Architectures such as ReAct \cite{yao2023react}, AutoGPT \cite{yang2023auto}, and LangChain \cite{chase2022langchain} have demonstrated autonomous decision-making via multi-step reasoning and API calls. ToolChain \cite{zhuang2023toolchain} and ToolLLM \cite{qin2023toolllm} extend this by integrating dynamic tool execution. However, these systems are mostly evaluated in highly simplistic settings, limiting their applicability to real-world operational workflows.

\textbf{Instruction Following and Benchmarking.}
Instruction following ability has been benchmarked through datasets like SUPER-NATURAL-INSTRUCTIONS \cite{wang2022supernaturalinstructions}, AlpacaEval \cite{Li_AlpacaEval_An_Automatic_2023}, and FollowEval \cite{jing2023followeval}. More recent multi-step benchmarks like ComplexBench \cite{wen2024complexbench} and InFoBench \cite{qin2024infobench} explore instruction complexity, but typically use machine-formatted inputs that omit the ambiguity and variability of human-authored SOPs.

\textbf{Tool Use and API-Centric Benchmarks.}
Benchmarks like Gorilla \cite{patil2023gorilla}, API-Bank \cite{li-etal-2023-api} and BENCHAGENTS ~\cite{butt2024benchagents} assess tool invocation and API usage capabilities. However, these focus on isolated tool interactions with minimal procedural context. In contrast, SOP execution involves coordinated tool use across dependent steps, often requiring error handling and state tracking.

\textbf{Agent Evaluation Frameworks.}
AgentBench \cite{liu2023agentbench} and PlanBench \cite{valmeekam2023planbench} offer general frameworks to evaluate LLM agents on planning and tool use. Yet, their task settings are narrow in scope and lack the procedural structure, conditional logic, and real-world ambiguity typical of industrial SOPs. Similarly, embodied agents are evaluated in ALFWorld \cite{shridhar2020alfworld} and BabyAI \cite{chevalier2018babyai}, but those environments (e.g., a kitchen) differ fundamentally from workflow automation tasks.

\textbf{SOP Automation and Business Process Modeling.}
Traditional business process automation relies on rule-based systems and formal process modeling languages \cite{dumas2018fundamentals, lindsay2003business, van2003workflow}. While effective, these require manual formalization of procedures, making them brittle and labor-intensive. Prior work on SOP formalization in organizational contexts \cite{de2009standard} has not yet translated into LLM-native solutions.
Recently, \cite{gschwind2025} proposes a novel LLM Classifier-Augmented Generation (CAG)
approach that translates procedures described in natural language into executable workflows. However, their target descriptions are rather short and limited to the domain of data integration. Most importantly, the dataset used to evaluate the approach is not publicly available. Recently, SOP-Maze \cite{wang2025sop} has published business operations SOPs, but they lack the tools or ground truth data for agent evaluation.

\textbf{How SOP-Bench Differs}
SOP-Bench addresses these gaps by offering (1) realistic SOP-style instructions, (2) diverse domain coverage, (3) structured APIs for execution, and (4) evaluation protocols grounded in
real-world complexity. Compared to benchmarks like Gorilla that isolate API usage, SOP-Bench evaluates agents on full workflows with inter-dependencies, ambiguity, and error handling requirements. It provides a more comprehensive testbed for assessing agent robustness in enterprise automation settings.

%% file: kdd_tex_files/sec03_benchmark_design.tex
\section{Generating Industry-grade SOPs}
\begin{figure*}[htb!]
 \centering 
  \includegraphics[width=0.8\textwidth]{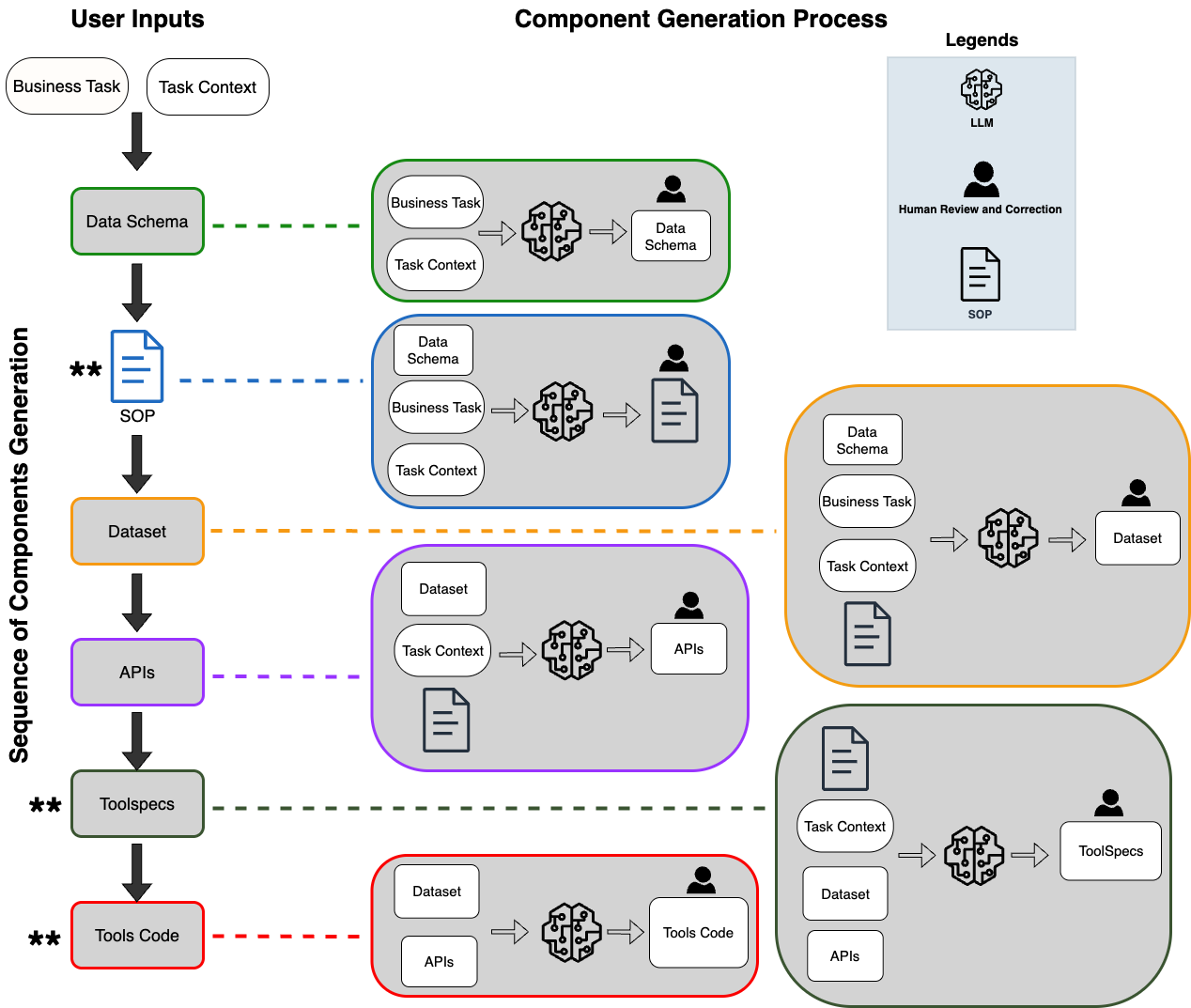}
    \caption{SOP-Bench Generation Workflow: Human experts author SOPs and validate all artifacts; AI generates tools, APIs, and datasets. \textbf{**} denotes post-generation complexity introduction.}
    \Description{Diagram illustrating the SOP-Bench generation workflow. Human experts create and validate SOPs, while an AI system generates tools, APIs, and datasets. Certain stages are annotated to indicate additional complexity introduced after generation.}    
    \label{fig:benchmark_factory_method}
\vspace{-0.25cm}
\end{figure*}

\setlength{\fboxrule}{1pt}
SOP-Bench is built on a novel human-AI collaborative framework where human domain experts author authentic SOPs while AI models generate supporting artifacts, all subsequently human-validated. This approach ensures procedural authenticity while maintaining reproducibility and scalability. The workflow employs a hierarchical prompting strategy using Anthropic's Claude 3.5 Sonnet v2 to refine human-authored SOPs for consistency and completeness, and to generate associated artifacts (tools, APIs, datasets) that capture real-world nuances: industry jargon, interdependent logic, implicit domain knowledge, ambiguous instructions, and cross-referenced content (Figure~\ref{fig:benchmark_factory_method}).

\subsection{Human-AI Collaborative Workflow}

\textbf{Human Expert Input:} Domain experts provide two critical inputs that define the benchmark scope:

\noindent
\textbf{Business Task}: The first version of an industrial use-case SOP

\noindent
\textbf{Task Context}: Procedural steps required to process requests successfully. For patient intake, this includes validating patient information, insurance details, medical history, and risk assessment.

{\small
\begin{itemize}
    \item \textbf{Business Task}: \texttt{SOP outlining steps for registering new patients}
    \item \textbf{Task Context}: \texttt{Validate insurance eligibility, prescription coverage, ...}
\end{itemize}
}

\definecolor{dataschema}{RGB}{23, 138, 21}
\colorlet{dataschema}{dataschema!70!white}
\subsubsection{\fcolorbox{dataschema}{white}{Dataset Schema} Generation (LLM-Assisted, Human-Validated)}
The first step is to create a structured dataset schema via one-shot prompting (Appendix \ref{appendix:sec:sop_data_schema_generation}). This schema specifies input parameters, decision points, compliance checks, and expected outcomes—acting as a semantic scaffold that minimizes hallucinations and ensures consistency. Human domain experts validate that the schema captures essential task components, including implicit knowledge and domain constraints. It uses the SOP and task context to generate the dataset schema. For example:
{\small
\begin{itemize}
    \item \textbf{Schema Entry}: \textbf{smoking\_status} -- \texttt{string}, \texttt{Choices = Never, Former, Current}
    \item \textbf{Without Schema} \textcolor{red}{\ding{55}}: \texttt{Calculate smoking risk using preferred methodology}
    \item \textbf{With Schema} \textcolor{green}{\ding{51}}: \texttt{Calculate smoking risk (Never, Former, Current)}
\end{itemize}
}

\definecolor{sop}{RGB}{32, 107, 193}
\colorlet{sop}{sop!70!white}
\subsubsection{\fcolorbox{sop}{white}{SOP Refinement} (Human-Authored, AI-Refined)}
Our workflow next uses the first version of the SOP, along with \textit{\textless Business Task, Task Context, Dataset Schema\textgreater} (Appendix \ref{appendix:sec:sop_document_generation}) to refine the SOP and make sure it is consistent with the data schema. The resulting SOP outlines procedure objectives, prerequisites, input requirements, detailed instructions, and decision logic in natural language. Human oversight validates domain standards, task logic, and corrects inconsistencies, ensuring SOPs are executable and contextually appropriate for industrial use.

\definecolor{dataset}{RGB}{246, 153, 18}
\subsubsection{\fcolorbox{dataset}{white}{Dataset} Generation (AI-Generated, Human-Validated)}
Using \textit{\textless Business Task, Task Context, Dataset Schema, SOP\textgreater} (Appendix~\ref{appendix:sec:sop_data_generation}), AI generates datasets representing full procedural context: structured inputs/outputs for every tool invocation, task parameters, decision pathways, and final outcomes. Human experts validate that each datapoint's logic is unambiguous, data types are appropriate, and the dataset accurately reflects SOP logic. The dataset includes challenging scenarios: positive/negative decision pathways, edge conditions, and task failure scenarios to test agent reasoning, compliance, error-handling, and tool selection.

\definecolor{api}{RGB}{153, 51, 255}
\colorlet{api}{api!70!white}
\definecolor{toolspec}{RGB}{58, 84, 49}
\colorlet{toolspec}{toolspec!70!white}
\subsubsection{\fcolorbox{api}{white}{API} \& \fcolorbox{toolspec}{white}{Tool Specs} Generation (AI-Generated, Human- Validated)}
Using \textit{\textless Task Context, Dataset, SOP\textgreater} (Appendix \ref{appendix:sec:sop_api_generation}), our workflow generates APIs specifying required inputs (e.g., \texttt{patient\_name}) and expected outputs (e.g., \texttt{patient\_id}), along with corresponding tool specifications (Figure~\ref{appendixfig:api_toolspec_example}). Human experts validate that APIs and ToolSpecs align with the dataset schema and access only relevant columns. Example:
{\small
\begin{itemize}
    \item \textbf{API Spec}: \textbf{getPatientID} -- retrieves patient identifier from name
    \item \textbf{Without API} \textcolor{red}{\ding{55}}: \texttt{def getPatientID(self): return \{\}}
    \item \textbf{With API} \textcolor{green}{\ding{51}}: \texttt{def getPatientID(self, patient\_name): return self.patientDB.lookupIDByName(patient\_name)}
\end{itemize}
}

\definecolor{toolscode}{RGB}{250, 0, 0}
\colorlet{toolscode}{toolscode!50!white}
\subsubsection{\fcolorbox{toolscode}{white}{Tools Code} Generation (AI-Generated, Human-Validated)}
Using \textit{\textless Dataset, APIs\textgreater} (Appendix \ref{appendix:sec:sop_api_code_generation}), Our workflow next generates executable tool code reflecting procedural logic. The dataset and API specs provide input-output contracts and data characteristics, ensuring code aligns with operational requirements. Human experts validate code correctness and consistency by executing them. 


\begin{table}[!t]
\vspace{-0.1cm}
\caption[Characteristics of Benchmark SOPs in SOP-Bench]{Characteristics of Benchmark SOPs in SOP-Bench. $n$ = number of tasks per SOP; $t$ = number of tools available; $\tau$ = token count, $\mathbb{C_H}$ = human-perceived complexity, average of three domain experts of three dimensions, ease of understanding, ambiguity or implicit domain knowledge and reasoning complexity, on a scale of 1 - 10; $\mathbb{C}_{llm}$ = complexity estimated by Claude 3.5 (scale 1-10, calibrated against Berkeley function calling tasks as baseline 2)}
\small
\centering
\begin{tabular}{@{}p{2.25cm}p{1.5cm}cccll@{}}
\toprule
\textbf{SOP Topic} & \textbf{Domain} & \textbf{$n$} & \textbf{$t$} & $\tau$ & $\mathbb{C}_{H}$ & $\mathbb{C}_{llm}$\\
\midrule
Content Flagging & Content Moderation & 226 & 4 & 850 & 7.67 & 9\\[0.5ex]
Customer Service & Support & 208 & 10 & 1457 & 9 & 8\\[0.5ex]
Dangerous Goods & Supply Chain & 327 & 5 & 775 & 5 & 7\\[0.5ex]
Aircraft Inspection & Transportation & 150 & 7 & 834 & 6.67 & 9 \\[0.5ex]
Email Intent & Retail Seller Mgmt & 122 & 4 & 766 & 5.33 & 7\\[0.5ex]
Know Your Business & Finance & 122 & 8 & 1359 & 8.33 & 9\\[0.5ex]
Patient Intake & Healthcare & 90 & 6 & 753 & 4.33 & 7\\[0.5ex]
Video Annotation & Autonomous Driving & 168 & 26 & 4492 & 9.67 & 10\\[0.5ex]
Video Classification & Media & 198 & 25 & 1249 & 8.33 & 9\\[0.5ex]
Warehouse Package Inspection & Logistics & 200 & 12 & 653 & 4.67 & 9\\[0.5ex]
Referral Abuse v1 & Trust \& Safety & 200 & 3 & 1502 & 6.33 & 7\\[0.5ex]
Referral Abuse v2 & Trust \& Safety & 200 & 6 & 2576 & 8.67 & 9\\[0.5ex]
Traffic Spoofing & Fraud Detection & 200 & 6 & 854 & 6.83 & 8\\[0.5ex]
\midrule
Average & & 185 & 9 & 1394 & 6.96 & 8.31\\
\midrule
Total & & 2411 & 122 & 18120 & \\
\bottomrule
\end{tabular}
\label{tab:sop_characteristics}
\vspace{-0.5cm}
\end{table}

\subsection{Human Authoring and Validation}
Domain experts perform five critical validation steps: (1) \textbf{SOP Authoring}: experts draft initial SOPs based on real industrial procedures, providing business context and task requirements; (2) \textbf{Schema Validation}: experts verify that AI-generated schemas capture essential task components, implicit domain knowledge, and compliance constraints; (3) \textbf{SOP Refinement}: experts review AI-refined SOPs for domain accuracy, logical consistency, and alignment with industrial standards; (4) \textbf{Dataset Validation}: experts validate {\textbf{each}} generated datapoint for logical correctness, appropriate data types, and coverage of edge cases; (5) \textbf{API \& Code Validation}: experts verify that generated APIs and executable code align with dataset schemas and operational requirements.

To quantify benchmark difficulty, three domain experts independently rated each SOP across three dimensions: ease of understanding, ambiguity/implicit knowledge requirements, and reasoning complexity (scale 1-10). These ratings, averaged to produce $\mathbb{C}_{H}$ in Table~\ref{tab:sop_characteristics}, reflect human-perceived complexity.


\subsection{SOP Descriptions}
SOP-Bench consists of 12 diverse SOPs instantiated in 2,411 tasks representing realistic industrial procedures. Each SOP tests different agent capabilities: complex decision-making, tool usage, and structured output generation across business modalities including classification (email intent), structured verification (business verification), industrial processes (aircraft inspection), and generic assignments (annotation, content flagging, customer service). Table~\ref{tab:sop_characteristics} shows task count ($n$), API count ($t$), SOP token count ($\tau$), and complexity ratings by three human domain experts ($\mathbb{C}_{H}$) and Claude 3.5 v2 ($\mathbb{C}_{llm}$).

\subsubsection{Controlled Complexity Variation}
To evaluate how procedural complexity affects agent performance, we include two versions of the referral abuse detection SOP. Version 1 (v1) implements a foundational fraud detection workflow with 3 tools and a simple three-outcome decision model based on current account indicators. Version 2 (v2) introduces significantly greater complexity through temporal pattern analysis, historical violation context, and risk-based enforcement, requiring 6 tools and producing 7 graduated enforcement actions. This controlled complexity variation enables systematic evaluation of how agents handle increasingly sophisticated reasoning requirements.

\subsubsection{Real-World Complexity in SOP-Bench}
To mirror authentic industrial conditions, human experts and AI collaboratively introduce controlled complexity at multiple levels across different SOPs. At the \textbf{SOP-level}, procedures contain branching logic, redundant details, tangential information, and ambiguous instructions that agents must filter—reflecting the reality that real-world SOPs are rarely perfectly structured. At the \textbf{tool-level}, we include semantically similar but functionally redundant tools to challenge agents' critical evaluation and tool selection capabilities. At the \textbf{data-level}, we introduce distractor variables that appear relevant but are not required for task completion.

Specific examples include: (1) \textbf{Video Annotation} and \textbf{Video Classification} contain 26 and 25 tools respectively, with multiple distractor tools that perform similar functions (e.g., different object detection methods, overlapping segmentation approaches), requiring agents to identify the correct tool sequence from numerous plausible alternatives; (2) \textbf{Patient Intake} includes distractor input variables such as patient demographic details and medical history fields that are provided but not necessary for certain decision pathways, testing whether agents can distinguish essential from supplementary information; (3) \textbf{Content Flagging} and \textbf{Customer Service} embed ambiguous conditional logic where threshold values and decision criteria require careful interpretation of SOP text rather than explicit enumeration. These interventions, validated through human oversight, rigorously assess tool selection behavior, information filtering, and reasoning under ambiguity. See Appendix \ref{appendix:sop_details} for detailed descriptions of all SOPs.

%% file: kdd_tex_files/sec04_agents_for_automating_sops.tex
\section{Agents for automating SOPs}

\begin{table*}[!htbp]
\vspace{-0.2cm}
\centering
\caption{Claude 4 Opus: Function Calling vs ReAct Agent Performance, ranked in descending order of ReAct TSR.}
\label{tab:claude4opus_comparison}
\small
\definecolor{lightgray}{gray}{0.9}
\begin{tabular}{l|cccc|cccc}
\toprule
\textbf{SOP Topic} & \multicolumn{4}{c|}{\textbf{Function Calling}} & \multicolumn{4}{c}{\textbf{ReAct}} \\
 & \textbf{ECR} & \textbf{C-TSR} & \textbf{TSR} & \textbf{Time (s)} & \textbf{ECR} & \textbf{C-TSR} & \textbf{TSR} & \textbf{Time (s)} \\
\midrule
\rowcolor{lightgray}
Patient Intake & 1.00 & 0.92 & 0.92 & 70.33 & 1.00 & 1.00 & 1.00 & 102.28 \\

Email Intent & 1.00 & 0.98 & 0.98 & 25.24 & 0.99 & 0.99 & 0.98 & 31.26 \\
\rowcolor{lightgray}
Referral Abuse Detection Easy & 1.00 & 0.95 & 0.95 & 53.99 & 1.00 & 0.97 & 0.97 & 66.45 \\

Aircraft Inspection & 1.00 & 0.45 & 0.45 & 83.11 & 1.00 & 0.97 & 0.97 & 87.46 \\
\rowcolor{lightgray}
Referral Abuse Detection Hard & 1.00 & 0.98 & 0.98 & 81.79 & 0.99 & 0.96 & 0.96 & 89.16 \\

Video Classification & 1.00 & 0.79 & 0.79 & 78.69 & 1.00 & 0.94 & 0.94 & 62.89 \\
\rowcolor{lightgray}
Dangerous Goods & 1.00 & 0.73 & 0.73 & 41.15 & 1.00 & 0.83 & 0.83 & 28.02 \\

Warehouse Package Inspection & 1.00 & 0.56 & 0.56 & 47.50 & 1.00 & 0.65 & 0.65 & 76.02 \\
\rowcolor{lightgray}
Traffic Spoofing Detection & 1.00 & 0.79 & 0.79 & 25.23 & 0.99 & 0.61 & 0.60 & 130.20 \\

Customer Service & 1.00 & 0.56 & 0.56 & 72.83 & 1.00 & 0.56 & 0.56 & 145.43 \\
\rowcolor{lightgray}
Video Annotation & 1.00 & 0.37 & 0.37 & 71.04 & 1.00 & 0.38 & 0.38 & 73.97 \\

Know Your Business & 0.99 & 0.44 & 0.43 & 159.84 & 1.00 & 0.31 & 0.31 & 110.50 \\
\rowcolor{lightgray}
Content Flagging & 1.00 & 0.36 & 0.36 & 49.93 & 0.99 & 0.28 & 0.27 & 138.08 \\
\midrule
\textbf{Average} & \textbf{1.00} & \textbf{0.68} & \textbf{0.68} & \textbf{66.21} & \textbf{1.00} & \textbf{0.73} & \textbf{0.72} & \textbf{87.82} \\
 \textbf{Standard Error} $\epsilon$ &  & \textbf{$\pm$0.07} & \textbf{$\pm$0.07} & \textbf{$\pm$9.57} & & \textbf{$\pm$0.08} & \textbf{$\pm$0.08} & \textbf{$\pm$10.30} \\
\bottomrule
\end{tabular}
\label{tab:evaluation}
\vspace{-0.2cm}
\end{table*}

To evaluate the effectiveness of LLM-based agents in executing industry grade SOPs, we implement two complementary agent architectures in SOP-Bench: (a) \textbf{Function Calling (FC) Agent} and (b) \textbf{ReAct Agent}. These implementations serve as baseline agents to demonstrate SOP-Bench's evaluation capabilities and are not intended to represent optimal architectures. We encourage the research community to develop and publish more sophisticated agent designs using SOP-Bench as the evaluation framework. Both baseline agents utilize LLM's native tool-calling interface to dynamically invoke tools/APIs during task execution. More details on these two agents are in Appendix ~\ref{appendix:technical_details_on_agents}.

\textbf{FC Agent} is a prompt-driven, lightweight execution engine leveraging native tool calling capabilities of LLMs. It accepts SOP text and task input, maintains a conversation loop with the LLM, and processes tool calls iteratively (up to 10 iterations). When the model triggers a tool use event, the agent executes the corresponding function via a ToolManager and returns results to the model. Final outputs are enclosed in XML tags for structured parsing.

\textbf{ReAct Agent} is built on a custom ReAct framework implementation~\cite{yao2023react}, designed for SOPs with complex branching logic and multi-step reasoning. It interleaves reasoning traces (Thoughts) with tool calls (Actions), feeding Observations back into the reasoning loop for up to 15 iterations. The agent enforces that at least one tool is executed before conclusions, maintaining comprehensive execution traces including intermediate steps and tool call records. More details on these two agents are in Appendix ~\ref{appendix:technical_details_on_agents}.

We evaluate both agents on SOP-Bench using standardized scripts that: (1) load SOP documents and structured task datasets; (2) convert each task into natural language instructions; and (3) execute agents while logging outputs, tool traces, and reasoning steps. To simulate API calls, each dataset includes precomputed inputs and outputs for every tool call, stored as columns. These mocks replace live APIs to enable stable, reproducible evaluation without runtime variability. In deployment, they would be swapped for real APIs. 


\subsection{Evaluation Methodology}
We evaluate FC-Agent and ReAct-Agent on SOP-Bench using three inputs: (1) Task, (2) SOP (instructions), and (3) ToolSpecs. The outputs include (1) \textit{intermediate tool outputs}, (2) \textit{final response}, and (3) \textit{trace of tool use and reasoning}. We define: $n$ (total number of tasks per SOP), $n_c$ (number of tasks per SOP marked complete by the agent), and $n_a$ (number of correctly completed tasks per SOP). Evaluation metrics are: (1) Execution Completion Rate ($ECR$): proportion of tasks the agent marked as complete, $ECR = \frac{n_c}{n}$; (2) Conditional Task Success Rate ($C$-$TSR$): fraction of completed tasks that match the ground truth, $C$-$TSR = \frac{n_a}{n_c}$; and (3) Task Success Rate ($TSR$): overall accuracy, $TSR = \frac{n_a}{n}$. Though this work reports only $ECR$ and $TSR$, future research should incorporate tailored metrics to better assess steps planning, tool mapping, tool calling accuracies, to better understand agent performance bottlenecks. We welcome the research community to contribute such agent evaluation metrics to SOP-Bench.

For all evaluations, we have kept the LLM parameters temperature at 0.5 and output token limits at 8000.

%% file: kdd_tex_files/sec05_results.tex
\section{Results}

\subsection{Results and Insights}
We evaluate the two agents, FC and ReAct, with 11 different LLMs in our experiments: Llama-3.3-70B-Instruct~\cite{llama3-1}, 
GPT-OSS-120B~\cite{openai2025gptoss}, DeepSeek-R1~\cite{deepseek2025r1}, Claude 3.5 Sonnet v2~\cite{anthropic2024claude35sonnet}, 
Claude 3.7 Sonnet~\cite{anthropic2025claude37sonnet}, Claude 4 Sonnet~\cite{anthropic2025claude4}, Claude 4 Opus~\cite{anthropic2025claude4}, 
Claude 4.1 Opus~\cite{anthropic2025claude4}, Claude Sonnet 4.5~\cite{anthropic2025claudesonnet45}, and Claude Opus 4.5~\cite{anthropic2025claudeopus45}. 
These models were selected based on availability within our organization and their known strong performance on agentic tasks, explaining the focus on larger, frontier models rather than smaller alternatives. Detailed experiment results of each SOP, Agent and LLM are in Appendix \ref{appendix:sop_details}. 

\subsubsection{LLM--Agent Architecture Must Be Task Dependent}

Our evaluation shows that no single agent architecture-model combination consistently dominates across SOP-Bench, highlighting the need for task-dependent architecture selection in SOP automation. 

Using Claude 4 Opus as a controlled comparison, Table~\ref{tab:claude4opus_comparison} contrasts Function-Calling (FC) and ReAct agents across all 13 SOPs. ReAct achieves a higher average Task Success Rate (TSR) than FC (72\% vs.\ 68\%), but at a significant latency cost: 87.82s $\pm$ 10.30s per task compared to 66.21s $\pm$ 9.57s for FC, a 33\% increase. Moreover, ReAct outperforms FC on only 8 of 13 SOPs, indicating that average gains obscure strong task-level variation. 

\textbf{Architecture--Model Co-Design Is Non-Monotonic.}
Performance does not scale monotonically with stronger models, particularly for reasoning-centric architectures. FC agents show modest, consistent improvements (Claude 3.5 v2: 65.8.2\% $\rightarrow$ Claude 4.5 Sonnet: 67.5\% TSR), while ReAct agents degrades: Claude 3.5 v2 achieves 65.4\% TSR, Claude 4 Sonnet improves to 68.7\%, but Claude 4.5 Sonnet drops to 63.3\% ($-2.1$ points). These results show that newer foundation models are not automatically better for existing agent architectures, and naive upgrades can silently reduce task success.

\textbf{Task Characteristics Dominate Aggregate Metrics.}
Aggregate averages mask clear task-specific reversals. Across SOPs, FC slightly outperforms ReAct on average (65.9\% vs.\ 61.1\% TSR), yet individual tasks exhibit distinct architectural preferences. Dangerous Goods classification shows near parity (FC: 69.1\%, ReAct: 69.8\%), while Customer Service reveals identical failure modes (FC: 47.8\%, ReAct: 47.7\%). Other SOPs strongly favor one architecture, with ReAct excelling in multi-step inspection workflows and FC performing better in deterministic, tool-driven procedures.

Overall, these findings demonstrate that agent architecture must be selected based on SOP structure, tool interaction patterns, and latency constraints rather than global averages. SOP-Bench enables systematic evaluation of these trade-offs, supporting principled architecture selection for real-world procedural workflows.

\begin{table}[htbp]
\centering
\caption{Top two performing model-agent combinations per SOP, ranked by Task Success Rate (TSR).}
\label{tab:top_two_tsr}
\footnotesize
\definecolor{lightgray}{gray}{0.9}
\begin{tabular}{p{1.4cm}p{1.2cm}p{0.6cm}c|p{1.2cm}p{0.6cm}c}
\toprule
\textbf{SOP Topic} & \multicolumn{3}{c|}{\textbf{Best}} & \multicolumn{3}{c}{\textbf{Second Best}} \\
\cline{2-7}
 & \textbf{Model} & \textbf{Agt} & \textbf{TSR} & \textbf{Model} & \textbf{Agt} & \textbf{TSR} \\
\midrule
\rowcolor{lightgray}
Patient Intake & Claude 4.1 Opus & ReAct & 1.00 & Claude 4.1 Opus & FC & 1.00 \\
Aircraft Inspection & Claude 3.7 Sonnet & ReAct & 0.99 & Claude 4.5 Opus & ReAct & 0.97 \\
\rowcolor{lightgray}
Email Intent & Claude 4 Sonnet & ReAct & 0.99 & Claude 4.1 Opus & FC & 0.98 \\
Referral Abuse Easy & Claude 3.5 v2 Sonnet & ReAct & 0.98 & Claude 3.5 v2 Sonnet & FC & 0.97 \\
\rowcolor{lightgray}
Referral Abuse Hard & Claude 4 Opus & FC & 0.98 & Claude 4.5 Opus & FC & 0.97 \\
Video Classification & Claude 4 Sonnet & FC & 0.95 & Claude 4 Sonnet & ReAct & 0.95 \\
\rowcolor{lightgray}
Dangerous Goods & Claude 4 Sonnet & FC & 0.87 & Claude 4.1 Opus & ReAct & 0.83 \\
Traffic Spoofing & Claude 4.5 Sonnet & FC & 0.86 & Claude 4 Sonnet & FC & 0.79 \\
\rowcolor{lightgray}
Customer Service & Llama 3.3 70B & ReAct & 0.79 & Claude 4.5 Opus & FC & 0.71 \\
Warehouse Inspection & Claude 4.1 Opus & ReAct & 0.69 & Claude 4 Opus & ReAct & 0.65 \\
\rowcolor{lightgray}
Content Flagging & Deepseek R1 & ReAct & 0.60 & Claude 3.5 v2 Sonnet & ReAct & 0.60 \\
Video Annotation & GPT OSS 120B & ReAct & 0.58 & Deepseek R1 & ReAct & 0.58 \\
\rowcolor{lightgray}
Know Your Business & Claude 4.5 Opus & ReAct & 0.58 & Claude 4.5 Haiku & FC & 0.57 \\
\bottomrule
\end{tabular}
\end{table}

\subsubsection{Domain Variance Exposes Generalization Limits}   

Agent performance on SOP-Bench varies substantially across procedural domains, exposing clear limits in cross-domain generalization. Table~\ref{tab:top_two_tsr} shows the top two model--agent combinations per SOP and confirms that no single configuration dominates across tasks. Best-case performance ranges from near-perfect accuracy on Patient Intake (100\% TSR) to only 57\% on Know Your Business, underscoring the necessity of domain-specific evaluation. Aggregated results in Table~\ref{tab:avg_tsr_by_model} further illustrate this variability: the FC agent achieves its highest average performance with Claude 4 Opus (68.2\% TSR), while ReAct performs best with Claude 4 Opus as well (73.8\% TSR). These rankings reflect a fixed agent configuration per model; further agent–model co-optimization could alter absolute performance but does not change the observed domain sensitivity.

Despite strong average performance for select models, domain-level results reveal a pronounced generalization gap. The easiest SOPs—Referral Abuse Detection (Easy) (94.3\% average TSR), Email Intent (88.7\%), and Patient Intake (88.1\%)—are over three times easier than the most challenging tasks, including Video Annotation (27.8\%), Aircraft Inspection (33.4\%), Content Flagging (38.0\%), and Warehouse Inspection (43.5\%), yielding a 3.4$\times$ performance range across SOPs.

Harder SOPs also exhibit greater performance variance across models. Video Annotation shows the highest standard deviation in TSR (13.8\%), indicating brittle and inconsistent agent behavior in long-horizon, tool-intensive workflows. Overall, these results demonstrate that success on one domain does not reliably transfer to others, validating SOP-Bench’s design goal of capturing orthogonal dimensions of procedural reasoning, tool use, and constraint adherence.

\begin{table}[!t]
\centering
\caption{Average Task Success Rate (TSR) by model and agent type, averaged across all SOPs. FC agent was only evaluated with Claude models due to its design for native tool calling. ReAct agent was evaluated across all models to compare reasoning capabilities.}
\label{tab:avg_tsr_by_model}
\footnotesize
\definecolor{lightgray}{gray}{0.9}
\begin{tabular}{lc||lc}
\toprule
\multicolumn{2}{c||}{{\textbf{ReAct Agent}}} & \multicolumn{2}{c}{{\textbf{FC Agent}}} \\
\cmidrule(lr){1-2} \cmidrule(lr){3-4}
\textbf{Model (rank)} & \textbf{TSR} & \textbf{Model (rank)} & \textbf{TSR} \\
\midrule
\rowcolor{lightgray}
Claude 4 Opus (1) & 0.724 & Claude 4 Opus (1) & 0.682 \\
Claude 4.1 Opus (2) & 0.694 & Claude 4 Sonnet (2) & 0.678 \\
\rowcolor{lightgray}
Claude 4 Sonnet (3) & 0.687 & Claude 4.5 Sonnet (3) & 0.675 \\
Claude 4.5 Opus (4) & 0.683 & Claude 4.5 Opus (4) & 0.667 \\
\rowcolor{lightgray}
Claude 3.5 v2 Sonnet (5) & 0.654 & Claude 4.1 Opus (5) & 0.664 \\
Claude 4.5 Sonnet (6) & 0.633 & Claude 3.5 v2 Sonnet (6) & 0.658 \\
\rowcolor{lightgray}
Llama 3.3 70B (7) & 0.565 & Claude 4.5 Haiku (7) & 0.624 \\
Deepseek R1 (8) & 0.557 & Claude 3.7 Sonnet (8) & 0.623 \\
\rowcolor{lightgray}
Claude 4.5 Haiku (9) & 0.522 & & \\
Claude 3.7 Sonnet (10) & 0.518 & & \\
\rowcolor{lightgray}
GPT OSS 120B (11) & 0.484 & & \\
\bottomrule
\end{tabular}
\vspace{-0.5cm}
\end{table}

\subsection{Ablation studies}

\subsubsection{Tool selection under registry bloat}  
To investigate the impact of tool registry size on agent performance, we conducted a controlled ablation study on the Video Annotation SOP using Claude Sonnet 4.5. We evaluated agent(react) performance under two conditions: (1) Bloated Registry: 26 tools (6 task-relevant + 20 distractor tools), and (2) Minimal Registry: 6 tools (task-relevant only). The minimal registry condition achieved 37\% TSR compared to 20.8\% TSR in the bloated condition, representing a 1.7x improvement. This suggests tool registry bloat could be a performance bottleneck: introducing 20 distractor tools reduced success rates by 16.2 percentage points despite agents having access to all necessary tools. Task-specific tool pruning may be necessary for practical deployment.

\vspace{-0.2cm}
\subsubsection{Controlled Complexity Variation in Referral Abuse Detection}

We conduct a controlled ablation on the Referral Abuse Detection SOP by comparing its \emph{Easy} and \emph{Hard} variants, which share identical task objectives, inputs, and tool interfaces but differ in logical complexity (Appendix~\ref{appendix:sop_details}). As shown in Table~\ref{tab:claude4opus_comparison}, Claude 4 Opus maintains similarly high TSR on both variants for both FC and ReAct agents, indicating robustness to increased procedural complexity. However, this accuracy comes at a clear reasoning cost: execution time increases by 51.5\% for the FC agent (53.99s $\rightarrow$ 81.79s) and by 34.2\% for the ReAct agent (66.45s $\rightarrow$ 89.16s) on the \emph{Hard} SOP. 
\vspace{-0.2cm}

\subsubsection{Long-Context Extraction, Not Conditional Reasoning, could limit Agent Performance}  

We observe variability in how context length and branching complexity affect performance. Long-context SOPs (e.g., Video Annotation: 4,492 tokens, 8 sequential steps, 2 branching decision points) show lower TSR (0.28), while high-branching SOPs (e.g., Referral Abuse Detection Hard: 2576 tokens, 28 sequential steps, 12 decision points) maintain high TSR (0.84). However, these SOPs differ in multiple dimensions (tool count, domain), preventing definitive conclusions. Further investigation with controlled SOP variants is needed to isolate the independent effects of context length versus branching complexity.
\vspace{-0.2cm}

%% file: kdd_tex_files/sec06_conclusion.tex
\section{Conclusions and Next Steps}
We present SOP-Bench, a comprehensive benchmark of 2,000+ executable tasks derived from human expert-authored SOPs across 12 industrial domains. SOP-Bench enables systematic evaluation of LLM agents under realistic procedural constraints, supporting principled investigation of agent architectures, model capabilities, and deployment strategies. Our baseline experiments across representative frontier models validate a central finding: \textbf{agent performance is highly task- and context-dependent, making rigorous evaluation on realistic SOPs essential prior to production deployment}.

Our results show that newer models do not necessarily yield better outcomes (e.g., Claude 4 family: 73.8\% vs.\ Claude 4.5 family: 58.8\% average TSR on ReAct), and that no single model–agent combination dominates across domains (TSR ranging from 57\% to 100\%). These findings illustrate how SOP-Bench can be used to analyze architecture–model interactions, surface performance bottlenecks, and inform deployment decisions without costly production experimentation. Rather than serving as a leaderboard, SOP-Bench provides an evaluation substrate for targeted scientific inquiry. It enables researchers to study procedural ambiguity, tool selection, and complexity scaling, while allowing practitioners to validate architecture choices against domain-specific requirements before deployment.

Looking ahead, we plan to extend SOP-Bench along several dimensions, while actively inviting community participation. We will introduce controlled variants of the same SOP with increasing complexity, incorporate multimodal instructions such as images and tables, and model hierarchical SOPs with nested structure and context switching. Beyond these extensions, we invite the community to leverage our human--AI collaborative methodology to author new SOPs, generate executable artifacts, and contribute domain-specific tasks back to SOP-Bench. This enables the benchmark to evolve organically, expand to new industries, and serve as a shared evaluation substrate for studying agent behavior.

We release the complete benchmark, baseline agents, and evaluation framework at \url{https://github.com/amazon-science/sop-bench}.

%% file: kdd_tex_files/appendix_gen-ai_use.tex
\section*{Use of Generative AI}
This work employed generative AI models as part of a human-AI collaborative framework for benchmark construction. Specifically, we used Anthropic's Claude 3.5 Sonnet v2 for the following purposes:

\textbf{Dataset Schema Generation:} Claude 3.5 Sonnet v2 generated structured dataset schemas from human-authored SOPs and task contexts using one-shot prompting. These schemas specified input parameters, decision points, and expected outcomes.

\textbf{SOP Refinement:} The model refined human-authored SOPs for consistency and completeness, ensuring alignment with dataset schemas while preserving domain-specific procedural logic.

\textbf{Artifact Generation:} Claude 3.5 Sonnet v2 generated supporting artifacts including: (1) synthetic datasets representing full procedural contexts with diverse decision pathways and edge cases, (2) API specifications defining tool inputs and outputs, (3) tool specifications for agent interaction, and (4) executable tool code implementing procedural logic.

\textbf{Complexity Assessment:} Claude 3.5 v2 provided complexity ratings for each SOP on a 1-10 scale, calibrated against Berkeley function calling tasks as baseline.

\textbf{Human Validation:} All AI-generated content underwent rigorous human validation by domain experts. Human experts: (1) authored all original SOPs based on real-world industrial procedures, (2) validated dataset schemas for completeness and domain accuracy, (3) reviewed and corrected SOP refinements for procedural correctness, (4) validated datasets for logical consistency and appropriate data types, (5) verified API and tool specifications aligned with dataset schemas, (6) executed and validated tool code for correctness, and (7) introduced controlled complexity to mirror real-world conditions. The human-AI collaborative approach ensured procedural authenticity while maintaining reproducibility and scalability.

\textbf{Evaluation:} The benchmark evaluation itself used 11 different LLMs (Claude 3.5 v2 Sonnet, Claude 3.7 Sonnet, Claude 4 Sonnet, Claude 4 Opus, Claude 4.1 Opus, Claude 4.5 Haiku, Claude 4.5 Sonnet, Claude 4.5 Opus, DeepSeek R1, GPT OSS 120B, Llama 3.3 70B) as the agents being evaluated, not as tools for analysis or writing.

\textbf{Manuscript Writing:} Claude 4.5 Sonnet was used to correct grammatical errors in the manuscript. All technical content, experimental design, analysis, and conclusions were authored by the human researchers.

\textbf{Main Image:} The main figure of the paper, which presents an overview of the Human–AI Collaborative workflow for SOP-Bench, was initially generated using ChatGPT, with the paper content provided as grounding material. Since the automatically generated figure did not fully capture the required structural and cosmetic refinements, the authors subsequently refined the figure manually to accurately represent the workflow, clearly delineate the Human and AI steps, and ensure faithful depiction of the overall process.

%% file: kdd_tex_files/appendix_agents.tex
\section{Technical Details on Agents}\label{appendix:technical_details_on_agents}

\subsection{FC Agent (Function-Calling Agent)}
This is a prompt-driven, lightweight execution engine leveraging the native tool calling capability of LLMs (e.g., Claude via AWS Bedrock). It accepts an \emph{SOP text} and a structured \emph{task input}, dynamically generates prompts using a parameterized template engine, and maintains a conversation loop with the LLM throughout execution. During agent initialization, it registers external tools by converting them from OpenAPI-style tool specifications into Bedrock's function-calling format. During execution, when the model triggers a \emph{tool\_use} event, the agent extracts the relevant tool name and parameters, executes the corresponding function via a domain-specific \emph{ToolManager}, and delivers the output to the model through a \emph{tool\_result} message. The agent implements an iterative conversation pattern with a maximum of 10 iterations, where each iteration processes tool calls sequentially until no further tool uses are detected. Final outputs are enclosed in XML-style tags (e.g., \emph{<final\_decision>}) and are parsed into structured formats for downstream evaluation. This architecture emphasizes traceability, minimal supervision, and broad applicability across various SOP domains. Note that this agent is incompatible with Amazon Nova models, which do not support native function calling.

\subsection{ReAct Agent}
Built on a custom implementation of the ReAct framework \cite{yao2023react}, this agent is designed to handle SOPs involving complex branching logic, intermediate reasoning, and decision pathways. It ingests the \emph{SOP text}, structured \emph{task input}, and tool specifications, similar to the FC Agent. Tools are dynamically discovered from the ToolManager and their descriptions are formatted with detailed parameter information including types, requirements, and allowed values. The ReAct Agent interleaves reasoning traces (\emph{Thoughts}) with explicit tool calls (\emph{Actions}), which are executed via the ToolManager. The resulting \emph{Observations} are fed back into the reasoning loop, enabling iterative planning and dynamic adaptation to SOP-specific control flows. The agent implements strict enforcement mechanisms to prevent premature final answers without tool usage, ensuring that at least one tool is executed before providing conclusions. Response parsing uses regular expressions to extract Thought, Action, Action Input, and Final Answer components, with robust error handling for malformed JSON inputs. This process continues for up to 15 iterations until a conclusive \emph{Final Answer} is reached, marked using tags such as \emph{<final\_decision>}. The agent maintains comprehensive execution traces including intermediate steps, tool call records with success status, and iteration counts. This agent architecture supports sophisticated workflows, enabling adaptive behavior in SOPs requiring fallback logic, verification steps, or multi-phase reasoning.

%% file: kdd_tex_files/appendix_sop_details.tex
\section{Details of the SOPs in SOP-Bench}
\label{appendix:sop_details}
In this section, we briefly describe the diverse industry use cases for which we generated synthetic data, and drawing parallels to real-world operational workflows.

\subsection{Aircraft Inspection}
This SOP establishes a framework for pre-flight airworthiness verification of aircrafts, covering multi-layered inspections like mechanical component checks, electrical system authentication, and maintenance record validation through seven distinct API tools. This SOP is valuable for evaluating LLM-powered agents, as it tests their ability to interpret technical parameters, manage complex multi-step workflows, and produce structured outputs for audit purposes. The complexity arises from its complex input structure and conditional branching (e.g., reporting inspection mismatches where found). It also has cross-domain applicability in industries like automotive manufacturing, healthcare equipment maintenance, and industrial machinery inspection, where precise validation and discrepancy reporting are critical.

\begin{table}[h]
\centering
\caption{Results for Aircraft Inspection SOP}
\label{tab:sop_aircraft inspection}
\begin{tabular}{llrr}
\toprule
\textbf{LLM Model} & \textbf{Agent} & \textbf{TSR} & \textbf{Exec Time (s)} \\
\midrule
Claude 3.7 Sonnet & ReAct & 0.990 & 23.04 \\
Claude 4.5 Opus & ReAct & 0.970 & 31.78 \\
Claude 4 Opus & ReAct & 0.970 & 87.46 \\
Claude 4.1 Opus & ReAct & 0.960 & 77.96 \\
Claude 4 Sonnet & ReAct & 0.880 & 23.74 \\
Claude 4.5 Haiku & ReAct & 0.870 & 16.26 \\
Claude 4.5 Sonnet & ReAct & 0.720 & 30.89 \\
Llama 3.3 70B & ReAct & 0.600 & 10.48 \\
Claude 4 Sonnet & Function Calling & 0.540 & 25.67 \\
Claude 3.7 Sonnet & Function Calling & 0.530 & 28.21 \\
Claude 3.5 v2 Sonnet & Function Calling & 0.510 & 25.97 \\
Claude 4.5 Sonnet & Function Calling & 0.460 & 32.89 \\
Claude 4 Opus & Function Calling & 0.450 & 83.11 \\
Claude 4.5 Haiku & Function Calling & 0.440 & 18.77 \\
Claude 4.1 Opus & Function Calling & 0.390 & 81.91 \\
Claude 4.5 Opus & Function Calling & 0.230 & 30.41 \\
Claude 3.5 v2 Sonnet & ReAct & 0.130 & 26.89 \\
\bottomrule
\end{tabular}
\end{table}

\subsection{Content Flagging}
This SOP models a content moderation workflow designed to evaluate flagged user-generated content through a combination of bot detection, trust scoring and violation severity assessment. The SOP mimics a workflow in use cases such as fraud detection in fintech or abuse prevention in social platforms, where automated AI and software systems have scored the content and the results of the upstream system are available as input to the human agent. The SOP integrates behavioral heuristics (e.g., captcha interaction, device fingerprinting), geolocation risk analysis, and multi-dimensional scoring frameworks such as the Bot Probability Index (BPI), User Trust Score, and Content Severity Index, etc. The process culminates in a final decision—ranging from content removal to user banning—based on a rule-based decision matrix. Its complexity arises from the interplay of user behavior history, system signals, and threshold-based branching logic, reflecting the nuanced reasoning required in high-stakes moderation pipelines.

This SOP is highly relevant because it simulates a structured yet non-trivial decision-making process involving conditional logic, multi-step tool usage, and real-world signal fusion. It relies on four distinct APIs (tools) that mirror common platform trust and safety operations, such as bot detection, trust scoring, severity assessment, and final content dispositioning. The evaluation criteria require agents to correctly compute decision outcomes and generate structured audit trails, offering a clear and rigorous way to assess agent planning accuracy, execution fidelity, and interpretability.

\begin{table}[h]
\centering
\caption{Results for Content Flagging SOP}
\label{tab:sop_content_flagging}
\begin{tabular}{llrr}
\toprule
\textbf{LLM Model} & \textbf{Agent} & \textbf{TSR} & \textbf{Exec Time (s)} \\
\midrule
Deepseek R1 & ReAct & 0.600 & 3.74 \\
Claude 3.5 v2 Sonnet & ReAct & 0.600 & 21.76 \\
Claude 3.5 v2 Sonnet & Function Calling & 0.570 & 17.45 \\
GPT OSS 120B & ReAct & 0.430 & 2.67 \\
Claude 4 Sonnet & ReAct & 0.390 & 25.42 \\
Claude 4.1 Opus & Function Calling & 0.390 & 50.04 \\
Claude 4.5 Haiku & Function Calling & 0.380 & 11.46 \\
Claude 4.5 Opus & Function Calling & 0.370 & 21.71 \\
Claude 4.5 Opus & ReAct & 0.370 & 38.69 \\
Claude 4.5 Sonnet & ReAct & 0.370 & 70.52 \\
Claude 4 Opus & Function Calling & 0.360 & 49.93 \\
Claude 4 Sonnet & Function Calling & 0.350 & 12.97 \\
Claude 3.7 Sonnet & Function Calling & 0.330 & 21.23 \\
Claude 4.5 Sonnet & Function Calling & 0.330 & 21.23 \\
Llama 3.3 70B & ReAct & 0.320 & 5.55 \\
Claude 4.1 Opus & ReAct & 0.290 & 131.01 \\
Claude 4 Opus & ReAct & 0.270 & 138.08 \\
Claude 4.5 Haiku & ReAct & 0.170 & 32.77 \\
Claude 3.7 Sonnet & ReAct & N/A & 33.92 \\
\bottomrule
\end{tabular}
\end{table}

\subsection{Customer Service}
This SOP defines an offline workflow for diagnosing and resolving customer-reported service issues based solely on initial inputs—such as an account ID and issue description—without real-time customer interaction. It is designed for backend support scenarios where resolution relies entirely on internal system data, logs, and historical telemetry, as commonly seen in telecommunications and infrastructure-based services. The procedure covers authentication validation, account eligibility checks, outage detection, diagnostic testing, troubleshooting, and escalation, using predefined thresholds for key metrics (e.g., latency, jitter, bandwidth). Traceability is ensured through session tokens, ticket IDs, timestamps, and audit logs.

This SOP is particularly suited for evaluating AI or human agents in semi-autonomous support systems, where success depends on accurate decision-making, conditional branching, and proper documentation. It simulates a realistic diagnostic environment through a unified API-like framework. Agents must interpret system signals (e.g., suspension status, outage presence, signal quality) and act accordingly. The final output is a structured JSON-based Resolution Summary Document (RSD), encoding the complete decision trail for reproducibility, auditability, and downstream integration.

\begin{table}[h]
\centering
\caption{Results for Customer Service SOP}
\label{tab:sop_customer_service}
\begin{tabular}{llrr}
\toprule
\textbf{LLM Model} & \textbf{Agent} & \textbf{TSR} & \textbf{Exec Time (s)} \\
\midrule
Llama 3.3 70B & ReAct & 0.790 & 13.70 \\
Claude 4.5 Opus & Function Calling & 0.710 & 30.57 \\
Claude 4 Sonnet & Function Calling & 0.630 & 19.69 \\
Claude 4.5 Sonnet & Function Calling & 0.630 & 35.11 \\
Claude 4 Sonnet & ReAct & 0.620 & 26.32 \\
Claude 3.5 v2 Sonnet & ReAct & 0.610 & 23.16 \\
Claude 3.7 Sonnet & Function Calling & 0.570 & 22.82 \\
Claude 4 Opus & Function Calling & 0.560 & 72.83 \\
Claude 4 Opus & ReAct & 0.560 & 145.43 \\
Claude 4.5 Opus & ReAct & 0.520 & 47.08 \\
Claude 4.1 Opus & ReAct & 0.470 & 67.48 \\
Claude 4.1 Opus & Function Calling & 0.470 & 71.13 \\
Claude 4.5 Sonnet & ReAct & 0.460 & 46.37 \\
Deepseek R1 & ReAct & 0.370 & 78.78 \\
GPT OSS 120B & ReAct & 0.340 & 15.88 \\
Claude 4.5 Haiku & ReAct & 0.310 & 53.68 \\
Claude 4.5 Haiku & Function Calling & 0.280 & 15.98 \\
Claude 3.5 v2 Sonnet & Function Calling & 0.220 & 21.16 \\
Claude 3.7 Sonnet & ReAct & 0.080 & 29.02 \\
\bottomrule
\end{tabular}
\end{table}

\subsection{Dangerous Goods Classification}
This benchmark includes a sophisticated SOP for dangerous goods classification in supply chain operations. The procedure integrates multiple data sources (Safety Data Sheets, Handling Guidelines, Transportation Requirements, and Disposal Guidelines) through a structured scoring system, where each component contributes a severity score (1-5) to determine the final hazard classification. The SOP incorporates practical considerations such as missing data handling, validation checks, and API integrations, culminating in a final classification system (Classes A through D) based on cumulative hazard scores. This example demonstrates the intricate decision-making processes and interdependencies typical in industrial settings, requiring AI agents to handle document parsing, numerical computations, conditional logic, and structured output generation - capabilities that go far beyond simple task execution.

Similar workflow patterns and complexity are found across various industrial applications, such as pharmaceutical quality control processes, aircraft pre-flight safety checks, food safety certifications, and industrial equipment maintenance prioritization. These use-cases share key characteristics including multi-source data integration, quantitative scoring systems, complex decision trees, comprehensive documentation needs, and API integrations, demonstrating how our benchmark represents realistic challenges faced in industrial automation.

\begin{table}[h]
\centering
\caption{Results for Dangerous Goods SOP}
\label{tab:sop_dangerous_goods}
\begin{tabular}{llrr}
\toprule
\textbf{LLM Model} & \textbf{Agent} & \textbf{TSR} & \textbf{Exec Time (s)} \\
\midrule
Claude 4 Sonnet & Function Calling & 0.870 & 13.11 \\
Claude 4 Opus & ReAct & 0.830 & 28.02 \\
Claude 4.1 Opus & ReAct & 0.830 & 31.35 \\
Claude 3.5 v2 Sonnet & ReAct & 0.810 & 6.95 \\
Claude 3.7 Sonnet & ReAct & 0.790 & 22.08 \\
Claude 3.5 v2 Sonnet & Function Calling & 0.760 & 13.02 \\
Claude 4.5 Opus & ReAct & 0.760 & 22.02 \\
Claude 3.7 Sonnet & Function Calling & 0.740 & 11.83 \\
Claude 4 Sonnet & ReAct & 0.730 & 14.73 \\
Claude 4.5 Sonnet & ReAct & 0.730 & 19.48 \\
Claude 4 Opus & Function Calling & 0.730 & 41.15 \\
Claude 4.1 Opus & Function Calling & 0.730 & 48.91 \\
Claude 4.5 Opus & Function Calling & 0.690 & 13.95 \\
Deepseek R1 & ReAct & 0.680 & 13.06 \\
GPT OSS 120B & ReAct & 0.640 & 5.35 \\
Claude 4.5 Sonnet & Function Calling & 0.640 & 15.78 \\
Llama 3.3 70B & ReAct & 0.560 & 60.13 \\
Claude 4.5 Haiku & ReAct & 0.330 & 20.18 \\
Claude 4.5 Haiku & Function Calling & 0.280 & 10.25 \\
\bottomrule
\end{tabular}
\end{table}

\subsection{Email intent}
This SOP is designed for processing seller appeals in e-commerce platforms, representing the complexity of modern customer communication systems. The procedure systematically handles email classification and response determination through a structured workflow that integrates multiple data sources and API endpoints. The SOP instructs categorizing seller communications into distinct categories (pricing concerns, description modifications, listing status inquiries, or general questions) and validate these against product databases, pricing systems, and inventory management through specific API calls. The procedure incorporates practical considerations such as data validation, error handling, and compliance documentation, culminating in standardized XML-formatted outputs with clear action items. This example demonstrates the intricate decision-making processes typical in modern business operations, requiring AI agents to handle NLU, multi-system data integration, and protocol-based response generation. Similar workflow patterns and complexity are found across various industries, such as customer support ticket systems, insurance claim processing, financial services communication handling, and travel booking support systems. These use-cases share key characteristics including intent classification needs, multiple API integrations, structured decision trees, standardized response protocols, and compliance documentation requirements.

\begin{table}[h]
\centering
\caption{Results for Email Intent SOP}
\label{tab:sop_email_intent}
\begin{tabular}{llrr}
\toprule
\textbf{LLM Model} & \textbf{Agent} & \textbf{TSR} & \textbf{Exec Time (s)} \\
\midrule
Claude 4 Sonnet & ReAct & 0.990 & 10.01 \\
Claude 4.5 Sonnet & ReAct & 0.980 & 10.64 \\
Claude 4.5 Opus & Function Calling & 0.980 & 12.04 \\
Claude 4 Opus & Function Calling & 0.980 & 25.24 \\
Claude 4.1 Opus & Function Calling & 0.980 & 28.60 \\
Claude 4 Opus & ReAct & 0.980 & 31.26 \\
Claude 4 Sonnet & Function Calling & 0.970 & 7.71 \\
Claude 4.1 Opus & ReAct & 0.970 & 30.89 \\
Claude 3.7 Sonnet & Function Calling & 0.957 & 9.78 \\
Claude 3.5 v2 Sonnet & ReAct & 0.950 & 7.08 \\
Claude 4.5 Sonnet & Function Calling & 0.930 & 10.62 \\
Claude 4.5 Haiku & ReAct & 0.920 & 7.42 \\
Claude 4.5 Opus & ReAct & 0.920 & 11.92 \\
Deepseek R1 & ReAct & 0.910 & 2.49 \\
Claude 4.5 Haiku & Function Calling & 0.900 & 6.05 \\
GPT OSS 120B & ReAct & 0.790 & 7.78 \\
Claude 3.7 Sonnet & ReAct & 0.790 & 22.08 \\
Claude 3.5 v2 Sonnet & Function Calling & 0.740 & 11.68 \\
Llama 3.3 70B & ReAct & 0.720 & 11.67 \\
\bottomrule
\end{tabular}
\end{table} 

\subsection{Know Your Business}
Verification of business entities is an essential component in the modern economy. This serves as the primary motivation for creating a comprehensive SOP focused on business verification that identifies high-risk entities before engaging in business relationships. The SOP outlines step-by-step guidelines on how to validate business details, registration, licenses, and tax identities, as well as ownership structures and sanctioned individuals. The SOP requires the agent to reason about completeness of business details and call relevant tools to fetch required data to aid in the verification process. Finally, the agent is required to either approve the business or escalate the case for human intervention.

\begin{table}[h]
\centering
\caption{Results for Know Your Business SOP}
\label{tab:sop_know_your_business}
\begin{tabular}{llrr}
\toprule
\textbf{LLM Model} & \textbf{Agent} & \textbf{TSR} & \textbf{Exec Time (s)} \\
\midrule
Claude 4.5 Opus & ReAct & 0.580 & 39.68 \\
Claude 4.5 Haiku & Function Calling & 0.570 & 25.80 \\
Claude 3.5 v2 Sonnet & Function Calling & 0.560 & 28.53 \\
Claude 3.7 Sonnet & Function Calling & 0.560 & 43.63 \\
Claude 3.5 v2 Sonnet & ReAct & 0.530 & 31.43 \\
Deepseek R1 & ReAct & 0.500 & 5.25 \\
Claude 4 Opus & Function Calling & 0.430 & 159.84 \\
GPT OSS 120B & ReAct & 0.420 & 2.10 \\
Claude 4.1 Opus & ReAct & 0.410 & 114.07 \\
Claude 4.5 Sonnet & Function Calling & 0.400 & 44.83 \\
Claude 4.5 Sonnet & ReAct & 0.390 & 57.03 \\
Llama 3.3 70B & ReAct & 0.370 & 3.57 \\
Claude 4.1 Opus & Function Calling & 0.350 & 105.25 \\
Claude 4 Sonnet & ReAct & 0.340 & 31.57 \\
Claude 4.5 Opus & Function Calling & 0.340 & 43.59 \\
Claude 4 Opus & ReAct & 0.310 & 110.50 \\
Claude 4.5 Haiku & ReAct & 0.240 & 53.73 \\
Claude 4 Sonnet & Function Calling & 0.170 & 28.18 \\
Claude 3.7 Sonnet & ReAct & N/A & 28.40 \\
\bottomrule
\end{tabular}
\end{table}

\subsection{Patient Intake}
This synthetic SOP models the end-to-end intake and registration process for new patients in clinical settings, incorporating insurance validation, prescription benefits verification, lifestyle and clinical risk stratification, and pharmacy network checks. Its complexity lies in the conditional decision pathways, cross-system interactions (e.g., mock prescription and pharmacy validations), and the integration of structured data inputs with algorithmic assessments like risk scoring. The SOP requires the agent to perform validation tasks using six APIs—each mapping to a distinct functional step—and to generate structured outputs that reflect real-world registration outcomes under healthcare compliance constraints.

As a benchmark dataset for evaluating LLM-based task planning agents, this SOP is highly relevant due to its realistic structure and deterministic yet context-sensitive branching logic. It simulates challenges found in broader industrial domains such as healthcare operations, insurance underwriting, and benefits management. Similar real-world workflows include hospital patient onboarding, insurance claims adjudication, and pharmacy eligibility reviews. The SOP's evaluation framework scores agents on whether they produce complete, correctly formatted outputs across six output fields and the final decision on patient registration, aligning well with goals of assessing LLM agent task planning, reliability, and tool execution precision.

\begin{table}[h]
\centering
\caption{Results for Patient Intake SOP}
\label{tab:sop_patient_intake}
\begin{tabular}{llrr}
\toprule
\textbf{LLM Model} & \textbf{Agent} & \textbf{TSR} & \textbf{Exec Time (s)} \\
\midrule
Llama 3.3 70B & ReAct & 1.000 & 6.58 \\
Claude 4.5 Haiku & Function Calling & 1.000 & 14.88 \\
Claude 4 Sonnet & Function Calling & 1.000 & 18.28 \\
Claude 4.5 Opus & Function Calling & 1.000 & 26.84 \\
Claude 4.5 Sonnet & Function Calling & 1.000 & 28.69 \\
Claude 4.5 Opus & ReAct & 1.000 & 52.84 \\
Claude 4.1 Opus & Function Calling & 1.000 & 77.13 \\
Claude 4.1 Opus & ReAct & 1.000 & 94.50 \\
Claude 4 Opus & ReAct & 1.000 & 102.28 \\
Claude 3.7 Sonnet & ReAct & 0.985 & 18.46 \\
Claude 4 Sonnet & ReAct & 0.985 & 21.13 \\
Claude 3.5 v2 Sonnet & Function Calling & 0.955 & 43.06 \\
Claude 4 Opus & Function Calling & 0.924 & 70.33 \\
Claude 4.5 Sonnet & ReAct & 0.833 & 43.89 \\
Claude 4.5 Haiku & ReAct & 0.561 & 24.98 \\
Claude 3.7 Sonnet & Function Calling & 0.515 & 12.93 \\
GPT OSS 120B & ReAct & 0.121 & 5.01 \\
Claude 3.5 v2 Sonnet & ReAct & N/A & N/A \\
\bottomrule
\end{tabular}
\end{table}

\subsection{Referral Abuse Detection (Version 1)}
This SOP establishes a framework for detecting and investigating referral abuse violations in subscription-based business environments, where malicious actors create fake accounts or use misleading promotional content to earn referral rewards fraudulently. The procedure guides investigators through a systematic multi-step process: account investigation (address verification, email analysis, website verification, business description review), account relations investigation (checking links to flagged accounts, analyzing login patterns and geographic consistency), and traffic analysis (reviewing revenue patterns, click-through rates, referral sources, and transaction history). The SOP employs a scoring-based decision framework where investigators calculate violation indicators across four categories—Abusive Account Creation, Misleading Ad Copy, Personal Orders (Related User), and No Violation—by counting boolean indicators from tool outputs (e.g., invalid addresses, suspicious email patterns, shared payment methods, connected accounts). The violation type is determined by the highest score meeting its threshold, with a priority ordering for tied scores.

This SOP is relevant for evaluating AI agents in fraud detection and trust \& safety operations, as it requires complex conditional reasoning, multi-source data integration, and structured decision-making under ambiguity. The procedure mirrors real-world referral fraud detection workflows used in e-commerce platforms, subscription services, and affiliate marketing programs. Agents must correctly interpret boolean fields, calculate multiple scoring metrics simultaneously, apply threshold-based logic, and handle edge cases such as inconclusive evidence. The final output is a structured enforcement action (Account Closure, No Action, or Inconclusive) based on the calculated violation type, requiring agents to demonstrate both analytical reasoning and precise rule execution.

\begin{table}[h]
\centering
\caption{Results for Referral Abuse Detection Easy SOP}
\label{tab:sop_referral_abuse_detection_easy}
\begin{tabular}{llrr}
\toprule
\textbf{LLM Model} & \textbf{Agent} & \textbf{TSR} & \textbf{Exec Time (s)} \\
\midrule
Claude 3.5 v2 Sonnet & ReAct & 0.980 & 11.04 \\
Claude 3.5 v2 Sonnet & Function Calling & 0.970 & 14.82 \\
Claude 4 Opus & ReAct & 0.970 & 66.45 \\
Claude 4 Sonnet & ReAct & 0.960 & 16.03 \\
Claude 3.7 Sonnet & Function Calling & 0.960 & 16.77 \\
Claude 4.5 Opus & ReAct & 0.960 & 19.17 \\
Claude 4.5 Haiku & Function Calling & 0.955 & 8.53 \\
Claude 4 Sonnet & Function Calling & 0.955 & 11.93 \\
Claude 4.5 Sonnet & Function Calling & 0.955 & 16.59 \\
Claude 4.1 Opus & Function Calling & 0.955 & 30.88 \\
Claude 4.1 Opus & ReAct & 0.955 & 54.18 \\
Claude 4 Opus & Function Calling & 0.954 & 53.99 \\
Claude 4.5 Opus & Function Calling & 0.950 & 17.45 \\
Claude 4.5 Sonnet & ReAct & 0.950 & 23.01 \\
Claude 4.5 Haiku & ReAct & 0.945 & 16.28 \\
Claude 3.7 Sonnet & ReAct & 0.695 & 22.45 \\
GPT OSS 120B & ReAct & 0.615 & 2.18 \\
Llama 3.3 70B & ReAct & 0.525 & 3.65 \\
Deepseek R1 & ReAct & 0.525 & 4.65 \\
\bottomrule
\end{tabular}
\end{table}

\subsection{Referral Abuse Detection (Version 2)}
This enhanced version of the referral abuse detection SOP introduces significantly greater complexity through temporal pattern analysis, historical violation context, and risk-based enforcement actions. Building upon the foundational framework of Version 1, this SOP adds three major dimensions: temporal fraud detection (analyzing registration bursts, off-hours activity patterns, and coordinated timing), historical context review (previous violations, warning history, account rehabilitation status, customer complaints), and a sophisticated risk severity classification system (CRITICAL, HIGH, MEDIUM, LOW) that determines graduated enforcement responses. The scoring framework is expanded with weighted indicators—registration bursts and coordinated account creation patterns receive double weight—and introduces a new Temporal Fraud Score category. The procedure requires agents to calculate risk severity based on multiple factors including revenue amount, violation history, complaint counts, and refund rates, then map the combination of violation type and risk severity to one of seven possible enforcement actions ranging from "Permanent Account Closure" to "Warning Issued" to "Manual Review Required."

The key difference between Version 1 and Version 2 lies in the decision complexity and contextual reasoning required. Version 1 uses a simpler three-outcome model (Account Closure, No Action, Inconclusive) based solely on current account indicators, while Version 2 implements a multi-tiered enforcement framework that considers account history, temporal patterns, and financial impact. Version 2 introduces 13 additional input fields (account age, previous violations, warning status, temporal patterns, customer complaints, refund rates), weighted scoring for high-confidence fraud indicators, and a two-stage decision process that first determines violation type then calculates risk severity. This mirrors real-world fraud operations where enforcement actions are proportional to offense severity and repeat offender status, requiring agents to demonstrate more sophisticated reasoning about risk assessment, historical context integration, and graduated response strategies.

\begin{table}[h]
\centering
\caption{Results for Referral Abuse Detection Hard SOP}
\label{tab:sop_referral_abuse_detection_hard}
\begin{tabular}{llrr}
\toprule
\textbf{LLM Model} & \textbf{Agent} & \textbf{TSR} & \textbf{Exec Time (s)} \\
\midrule
Claude 4 Opus & Function Calling & 0.980 & 81.79 \\
Claude 4.5 Opus & Function Calling & 0.975 & 22.94 \\
Claude 3.7 Sonnet & Function Calling & 0.970 & 21.81 \\
Claude 4.5 Opus & ReAct & 0.965 & 31.55 \\
Claude 4.1 Opus & ReAct & 0.965 & 81.41 \\
Claude 4 Opus & ReAct & 0.960 & 89.16 \\
Claude 4.1 Opus & Function Calling & 0.955 & 73.86 \\
Claude 3.5 v2 Sonnet & Function Calling & 0.950 & 21.25 \\
Claude 4.5 Sonnet & ReAct & 0.945 & 45.50 \\
Claude 4.5 Sonnet & Function Calling & 0.935 & 16.15 \\
Claude 4.5 Haiku & ReAct & 0.915 & N/A \\
Claude 3.5 v2 Sonnet & ReAct & 0.915 & 14.57 \\
Claude 4 Sonnet & ReAct & 0.915 & 21.88 \\
Claude 4 Sonnet & Function Calling & 0.895 & 16.94 \\
Claude 4.5 Haiku & Function Calling & 0.670 & 14.94 \\
GPT OSS 120B & ReAct & 0.185 & 1.11 \\
Claude 3.7 Sonnet & ReAct & 0.100 & 29.25 \\
Llama 3.3 70B & ReAct & 0.085 & 8.50 \\
Deepseek R1 & ReAct & 0.060 & 6.00 \\
\bottomrule
\end{tabular}
\end{table}

\subsection{Traffic Spoofing Detection}
This SOP establishes a comprehensive framework for detecting and investigating traffic spoofing violations in affiliate marketing environments. The procedure guides fraud analysts through systematic validation of partner accounts, traffic pattern analysis, and source authentication to identify fraudulent activities such as IP/referrer falsification, content manipulation, and unauthorized redirections. The SOP employs multiple scoring thresholds including Engagement Score Threshold (EST), Conversion Rate Authentication Protocol (CRAP), and device traffic distribution analysis to classify violations. Investigators must validate metrics such as user-to-order ratios (minimum 0.4 for legitimacy), engagement scores (critical if <1.0), conversion rates (suspicious if <0.05\%), bounce rates, visit durations, and unattributed click percentages. The procedure requires cross-referencing multiple data sources through the Traffic Attribution Validation Matrix (TAVM) and documenting evidence in the Partner Violation Documentation System (PVDS). Based on risk assessment algorithms, the SOP determines graduated enforcement actions ranging from "Account Closure" for high-risk violations to "Warning Issued" or "No Action" for lower-risk cases. This SOP is relevant for evaluating AI agents in fraud detection and compliance operations, as it requires multi-dimensional data analysis, threshold-based decision logic, evidence correlation across traffic sources, and risk-proportional enforcement determination—capabilities essential for automated trust and safety systems in digital advertising and affiliate marketing platforms.

\begin{table}[h]
\centering
\caption{Results for Traffic Spoofing Detection SOP}
\label{tab:sop_traffic_spoofing_detection}
\begin{tabular}{llrr}
\toprule
\textbf{LLM Model} & \textbf{Agent} & \textbf{TSR} & \textbf{Exec Time (s)} \\
\midrule
Claude 4.5 Sonnet & Function Calling & 0.860 & 28.62 \\
Claude 4 Sonnet & Function Calling & 0.790 & 17.94 \\
Claude 4 Opus & Function Calling & 0.785 & 25.23 \\
Claude 4.5 Haiku & Function Calling & 0.750 & 9.76 \\
Llama 3.3 70B & ReAct & 0.740 & 3.35 \\
Claude 4.1 Opus & Function Calling & 0.720 & 20.04 \\
Claude 3.5 v2 Sonnet & Function Calling & 0.690 & 16.09 \\
Claude 3.7 Sonnet & Function Calling & 0.635 & 25.86 \\
Deepseek R1 & ReAct & 0.625 & 5.77 \\
Claude 4 Opus & ReAct & 0.600 & 130.20 \\
GPT OSS 120B & ReAct & 0.455 & 1.76 \\
Claude 4.5 Sonnet & ReAct & 0.365 & 48.60 \\
Claude 4.5 Opus & Function Calling & 0.350 & 31.50 \\
Claude 4 Sonnet & ReAct & 0.345 & 34.05 \\
Claude 3.5 v2 Sonnet & ReAct & 0.330 & 21.97 \\
Claude 4.1 Opus & ReAct & 0.225 & 120.45 \\
Claude 4.5 Haiku & ReAct & 0.135 & 33.70 \\
Claude 4.5 Opus & ReAct & 0.120 & 52.83 \\
Claude 3.7 Sonnet & ReAct & 0.025 & 29.64 \\
\bottomrule
\end{tabular}
\end{table}

\subsection{Video Annotation for self driving}
As we rapidly move to a world powered by AI where autonomous vehicles navigate our streets, there is a need to produce high-quality datasets to train perception and navigation systems. This serves as the main motivation behind creating a SOP that outlines steps to detect objects and annotate videos. The SOP establishes input requirements (format, frame rate, resolution, color depth), comprehensive processes for object detection and segmentation using specialized tools, and strict quality control measures (spatial accuracy, temporal consistency, inter-annotator agreement). After processing, the agent must report whether the video was successfully processed, provide paths to the resulting object detection and segmentation files, and include human reviewer scores to validate annotation quality.

\begin{table}[h]
\centering
\caption{Results for Video Annotation SOP}
\label{tab:sop_video_annotation}
\begin{tabular}{llrr}
\toprule
\textbf{LLM Model} & \textbf{Agent} & \textbf{TSR} & \textbf{Exec Time (s)} \\
\midrule
GPT OSS 120B & ReAct & 0.584 & 3.16 \\
Deepseek R1 & ReAct & 0.576 & 5.66 \\
Claude 4.5 Opus & Function Calling & 0.520 & 80.50 \\
Claude 3.5 v2 Sonnet & ReAct & 0.469 & 25.68 \\
Llama 3.3 70B & ReAct & 0.376 & 3.92 \\
Claude 4 Opus & ReAct & 0.376 & 73.97 \\
Claude 4 Opus & Function Calling & 0.368 & 71.04 \\
Claude 4.1 Opus & Function Calling & 0.360 & 79.65 \\
Claude 4.1 Opus & ReAct & 0.320 & 62.87 \\
Claude 4 Sonnet & ReAct & 0.312 & 18.71 \\
Claude 4.5 Haiku & Function Calling & 0.312 & 20.79 \\
Claude 4.5 Haiku & ReAct & 0.312 & 23.80 \\
Claude 4.5 Opus & ReAct & 0.280 & 38.90 \\
Claude 3.7 Sonnet & ReAct & 0.240 & 37.67 \\
Claude 4.5 Sonnet & ReAct & 0.208 & 42.43 \\
Claude 3.5 v2 Sonnet & Function Calling & 0.176 & 30.78 \\
Claude 4 Sonnet & Function Calling & 0.160 & 32.52 \\
Claude 4.5 Sonnet & Function Calling & 0.136 & 36.08 \\
Claude 3.7 Sonnet & Function Calling & 0.048 & 28.66 \\
\bottomrule
\end{tabular}
\end{table}

\subsection{Video Flagging for Content Moderation}

In the online world of content consumption and social media, content moderation plays a critical role in ensuring user safety, maintaining community standards, and preventing the spread of harmful or misleading information. As platforms scale, the volume and variety of user-generated content increase exponentially, making manual moderation infeasible. This has led to the growing reliance on automated moderation systems powered by artificial intelligence and machine learning. This SOP establishes a comprehensive framework for the systematic classification, escalation, and moderation of user-generated video content on digital platforms. The Agent uses tools to fetch initial review metadata and uses reasoning to escalate the content to human moderator and based on their feedback, assigns moderation tags, content warning status and a final decision to ultimately keep or take down the video.

\begin{table}[h]
\centering
\caption{Results for Video Classification SOP}
\label{tab:sop_video_classification}
\begin{tabular}{llrr}
\toprule
\textbf{LLM Model} & \textbf{Agent} & \textbf{TSR} & \textbf{Exec Time (s)} \\
\midrule
Claude 4 Sonnet & Function Calling & 0.954 & 22.19 \\
Claude 4 Sonnet & ReAct & 0.954 & 22.19 \\
Deepseek R1 & ReAct & 0.949 & 5.59 \\
Claude 4.5 Sonnet & ReAct & 0.949 & 19.97 \\
Claude 4.5 Sonnet & Function Calling & 0.949 & 23.70 \\
Claude 4.1 Opus & ReAct & 0.944 & 60.01 \\
Claude 3.7 Sonnet & ReAct & 0.939 & 26.90 \\
Claude 4 Opus & ReAct & 0.939 & 62.89 \\
Claude 4.5 Haiku & Function Calling & 0.929 & 12.95 \\
GPT OSS 120B & ReAct & 0.924 & 3.82 \\
Claude 4.5 Haiku & ReAct & 0.924 & 20.70 \\
Llama 3.3 70B & ReAct & 0.914 & 12.37 \\
Claude 4.5 Opus & ReAct & 0.914 & 30.28 \\
Claude 4.5 Opus & Function Calling & 0.914 & 35.95 \\
Claude 3.5 v2 Sonnet & Function Calling & 0.898 & 28.12 \\
Claude 3.5 v2 Sonnet & ReAct & 0.893 & 26.54 \\
Claude 4.1 Opus & Function Calling & 0.807 & 82.61 \\
Claude 3.7 Sonnet & Function Calling & 0.792 & 24.04 \\
Claude 4 Opus & Function Calling & 0.787 & 78.69 \\
\bottomrule
\end{tabular}
\end{table}

\begin{figure*}[htbp]
 \centering
  \includegraphics[width=\textwidth]{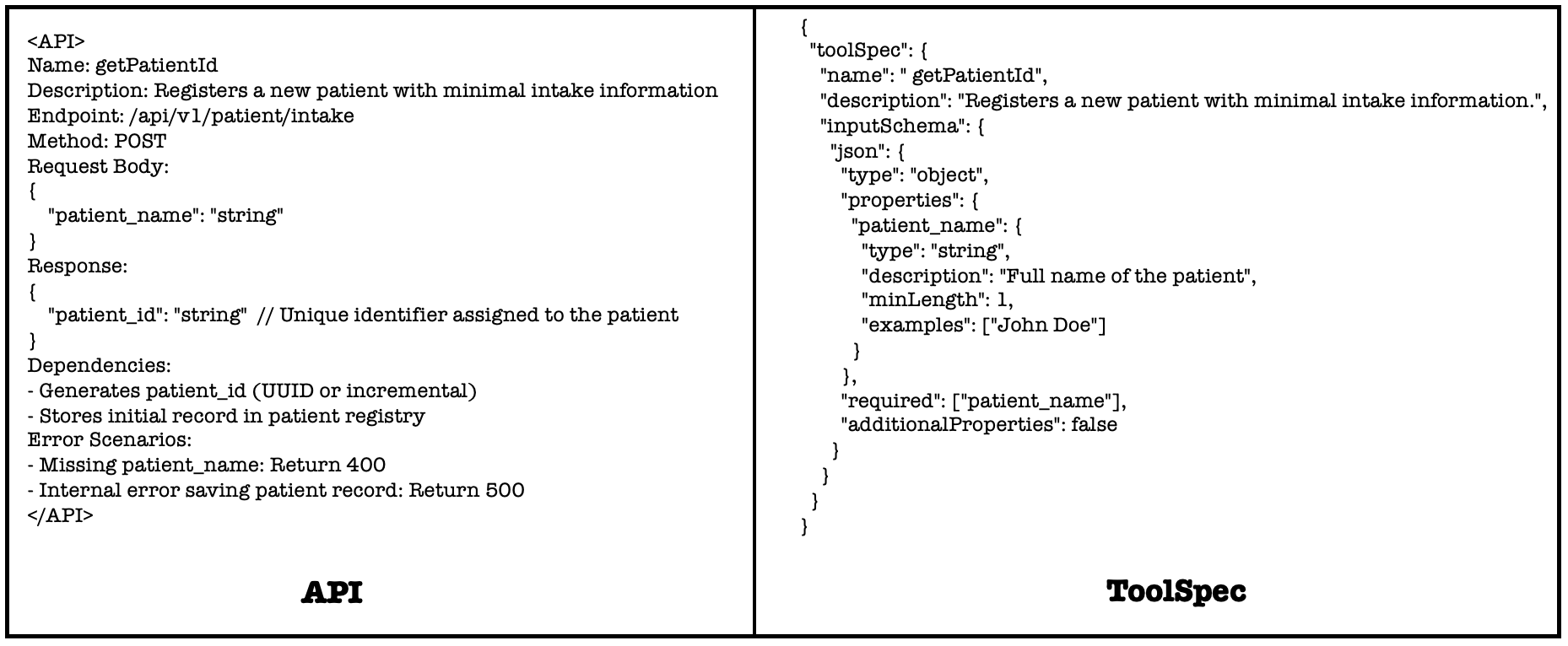}
    \caption{Sample API and ToolSpec generated for the SOP related to Patient Intake in Healthcare industry}
    \Description{Example screenshot showing a generated API specification and corresponding tool specification for a patient intake SOP in the healthcare domain.}
    \label{appendixfig:api_toolspec_example}
\vspace{-0.25cm}
\end{figure*}

\begin{figure*}[htbp]
 \centering
  \includegraphics[width=\textwidth]{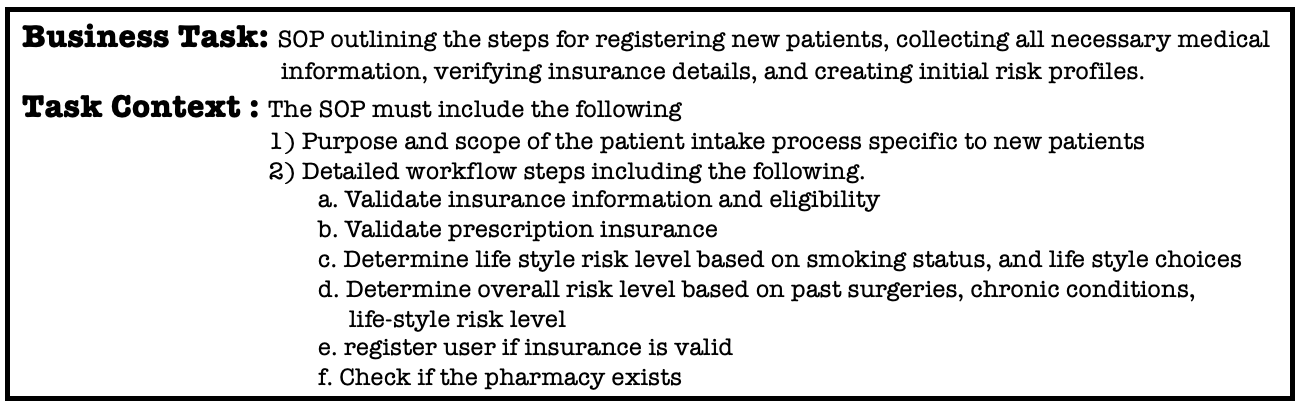}
    \caption{Detailed Sample of Business Task and Task Context for the Patient Intake SOP relevant to healthcare industry}
    \Description{Screenshot illustrating a detailed business task and its associated task context for a patient intake SOP in the healthcare domain, showing structured fields and procedural information.}
    \label{appendixfig:full_business_task_context}
\vspace{-0.25cm}
\end{figure*}

\subsection{Warehouse Package Inspection}
This SOP enhances the warehouse package inspection workflow by introducing a structured protocol for detecting and resolving shipment discrepancies. It reflects real-world operational needs in enterprise logistics, where issues like barcode mismatches, quantity variances, and damaged goods require swift, automated handling to maintain inventory integrity and vendor accountability.

The process begins with barcode validation using image comparison. If the barcode does not match the confirmed product ID, the item is immediately classified as “Wrong Item” and marked for return, bypassing further checks. Otherwise, the agent calculates quantity variance and classifies discrepancies based on predefined rules—such as overage, underage, cancelled quantity, or severe mismatch when the variance exceeds a set threshold.

Subsequent checks include verifying warehouse location and assessing package condition via image-based damage detection. Discrepancies like misdirected shipments or visible damage are accordingly classified. Once all issues are identified, the system calculates financial penalties using a chargeback matrix, applying formulas based on discrepancy type and unit cost.

The procedure concludes with structured outputs: a problem classification report, chargeback sheet, and resolution status update. Designed to simulate operational flows in high-volume retail logistics, this SOP challenges agents to integrate vision processing, rule-based logic, and structured reporting—providing a rigorous benchmark for LLM agents in industrial SOP execution.

\begin{table}[h]
\centering
\caption{Results for Warehouse Package Inspection SOP}
\label{tab:sop_warehouse_package_inspection}
\begin{tabular}{llrr}
\toprule
\textbf{LLM Model} & \textbf{Agent} & \textbf{TSR} & \textbf{Exec Time (s)} \\
\midrule
Claude 4.1 Opus & ReAct & 0.687 & 71.90 \\
Claude 4.5 Opus & Function Calling & 0.647 & 18.54 \\
Claude 4 Opus & ReAct & 0.647 & 76.02 \\
Claude 4.5 Haiku & Function Calling & 0.640 & 9.23 \\
Claude 3.5 v2 Sonnet & ReAct & 0.627 & 34.04 \\
Claude 3.5 v2 Sonnet & Function Calling & 0.560 & 18.19 \\
Claude 4 Opus & Function Calling & 0.560 & 47.50 \\
Claude 4.5 Sonnet & Function Calling & 0.553 & 16.50 \\
Claude 4 Sonnet & Function Calling & 0.527 & 15.22 \\
Claude 4.1 Opus & Function Calling & 0.527 & 49.67 \\
Claude 4.5 Opus & ReAct & 0.513 & 19.48 \\
Claude 4 Sonnet & ReAct & 0.513 & 26.19 \\
Claude 3.7 Sonnet & Function Calling & 0.493 & 17.68 \\
Llama 3.3 70B & ReAct & 0.346 & 7.91 \\
Deepseek R1 & ReAct & 0.333 & 5.66 \\
Claude 4.5 Sonnet & ReAct & 0.327 & 40.66 \\
GPT OSS 120B & ReAct & 0.300 & 3.31 \\
Claude 4.5 Haiku & ReAct & 0.147 & 27.08 \\
Claude 3.7 Sonnet & ReAct & 0.067 & 36.53 \\
\bottomrule
\end{tabular}
\end{table}

%% file: kdd_tex_files/appendix_figures.tex



%% file: kdd_tex_files/appendix_prompt.tex
\section{Prompt Templates}

\subsection{SOP Document Generation Prompt}
\label{appendix:sec:sop_document_generation}
\begin{lstlisting}[language=yaml, caption={Prompt Template YAML for converting user-provided task title and context to SOP}, breaklines=true, breakindent=0pt, columns=flexible, label={lst:sop_prompt}]
prompts:
  - name: complex_sop_generator
    parameters: [sop_title, additional_context, sample_schema]
    text: |
      You are an expert AI system specialized in creating highly detailed Standard Operating Procedures (SOPs) for complex, technical, and regulated environments. Your task is to generate an SOP that is rich in domain-specific terminology, intricate logic, and advanced structure-requiring substantial subject-matter expertise to fully comprehend.

      Here is the title for the SOP you need to create:
      <sop_title>
      {{sop_title}}
      </sop_title>

      {% if additional_context != "" %}
      Here is additional context that you must incorporate into the SOP generation process:
      <additional_context>
      {{additional_context}}
      </additional_context>
      {% endif %}

      {% if sample_schema != "" %}
      Here is a schema for the SOP that you should use to guide your creation process:
      <sample_schema>
      {{sample_schema}}
      </sample_schema>
      {% endif %}

      Before generating the SOP, you must thoroughly analyze the title, additional context  and sample schema to plan the most sophisticated and complex approach possible. Conduct this analysis and planning phase inside <sop_analysis> tags. Follow these steps:

      1. Deconstruct the title and additional context into key components.
      2. Identify the specific industry or field the SOP pertains to.
      3. Enumerate potential regulatory requirements relevant to this SOP.
      4. List out key technical terms that will be used in the SOP.
      5. Brainstorm complex technical details for each section of the SOP.
      6. Outline interdependencies between different sections of the SOP.
      7. Consider potential complications, edge cases, and rare scenarios that should be addressed.
      8. Identify specific compliance issues and how they will be addressed in the SOP.

      Use industry-specific jargon and complex terminology throughout the planning process. Consider industry-specific nuances, potential regulatory requirements, and intricate technical details that could be incorporated. This planning phase should be extensive and detailed.

      After completing the planning phase, generate the SOP using the following structure:

      <SOP>
      1. Purpose
      [Provide a clear, technical statement of the purpose of SOP]

      2. Scope
      [Define the exact coverage of the SOP and specify to whom it applies]

      3. Definitions
      [List and define relevant terms, ensuring they are technical and field-specific]

      4. Input
      [Enumerate required inputs, materials, or information needed to initiate the procedure]

      5. Main Procedure
      5.1 [First main step]
        5.1.1 [Substep with detailed explanation]
        5.1.2 [Substep with detailed explanation]
      5.2 [Second main step]
        5.2.1 [Substep with detailed explanation]
        5.2.2 [Substep with detailed explanation]
      [Continue with all necessary steps and substeps, ensuring each is explained in detail]

      6. Output
      [Describe the expected results or deliverables upon completion of the procedure]
      </SOP>

      Requirements for SOP Generation:

      1. Use advanced industry jargon, long-form technical sentences, and assume your audience has deep domain expertise.
      2. Ensure that the language used throughout the SOP is sophisticated and technical, requiring a high level of expertise to fully comprehend.
      3. Include cross-references between different sections to create a web of interconnected information.
      4. In the Definitions section, include complex terms that will be used throughout the SOP, ensuring they are technical and field-specific.
      5. Use the information provided in the additional context to construct the SOP.
      6. Do not use bullet points or lists. Instead, express each item in fully elaborated prose, forming complete paragraphs. Each requirement or specification must be described with technical detail, rationale, and interconnections, mimicking the tone of formal documentation or academic writing.
      7. At each stage, explicitly specify quantitative thresholds and conditions that guide decisions (e.g., scoring cutoffs, routing rules, escalation triggers).
      8. The SOP should be self-sustaining, all necessary instructions to complete the task must be explicitly mentioned in the SOP
      9. If a sample schema is provided, use the information exhaustively to design thresholds, success and failure logic to guide decision-making in the SOP, increasing its complexity.
      10. For each section, use the <sop_analysis> tags to think and reason about ways to increase complexity and depth. Consider potential pitfalls, rare scenarios, and intricate details that might not be immediately obvious.
      11. Generate the entire SOP in one continuous pass without pausing, asking for confirmation, or requesting interaction. Do not ask if you should continue. Output the full SOP all at once within the <SOP> </SOP> tags.

      Remember to make each step and substep in the Main Procedure detailed and clearly explained, while maintaining the required level of complexity and technical depth. This is crucial for the effectiveness of the SOP in its intended environment.
\end{lstlisting}

\subsection{Data Schema Generation Prompt}
\label{appendix:sec:sop_data_schema_generation}
\begin{lstlisting}[language=yaml, caption={Prompt Template YAML for converting user-provided task title, task context and a one-shot demonstration of a dataset schema to Synthetic Dataset Schema}, breaklines=true, breakindent=0pt, columns=flexible, label={lst:dataschema_prompt}]
prompts:
    - name: generate_dataset_schema
    parameters: [sop_title, additional_context]
    text: |
      You are an expert AI assistant that specializes in generating synthetic dataset schemas based on Standard Operating Procedures (SOPs). 
      You will use the  Standard operating procedure(SOP) text to generate a dataset schema.
      
      First, carefully read through the following SOP title:

      <sop_title>
        {{sop_title}}
      </sop_title>

      {% if additional_context != "" %} 
      Here is additional context you must adhere to for the SOP generation process: 
        <additional_context>
        {{additional_context}}
        </additional_context>
      {% endif %}

      You must structure the output using this format:
      <schema>
        video_id - string, unique identifier for the video (e.g., "vid_00123").
        video_path - string, file path or URI of the video (e.g., "/data/videos/vid_00123.mp4").
        upload_timestamp - datetime, when the video was uploaded (e.g., "2025-04-18T15:32:10Z").
        uploader_id - string, anonymized user identifier (e.g., "user_487").
        video_language - string, detected or declared language of spoken/audio content (e.g., "en", "es").
        region - string, region associated with the video or uploader (e.g., "US", "IN").
        metadata_tags - list[string], keywords or tags extracted from video metadata (e.g., ["protest", "crowd", "chanting"]).
        format_validated - boolean, whether the video passed format checks (e.g., True).
        duration_seconds - integer, total length of the video in seconds (e.g., 312).
      </schema>

      Now, Generate a dataset schema within <schema></schema> tags for the given SOP title using the additional context .
      <requirements>
      1. Define the schema of the dataset, including each field:
        - Name
        - Type (e.g., string, integer, float, boolean, datetime, enum, list, etc.)
        - Description (what the field represents)
        - Example value
      2. Include data ranges and choices wherever applicable.
      </requirements>
\end{lstlisting}

\subsection{Dataset Generation Prompt}
\label{appendix:sec:sop_data_generation}
\begin{lstlisting}[language=yaml, caption={Prompt Template YAML for converting user-provided task title, task context and dataset schema to Synthetic Dataset}, breaklines=true, breakindent=0pt, columns=flexible, label={lst:dataset_prompt}]
prompts:
    - name: generate_dataset_csv
    parameters: [sop_title, sop_file_contents, additional_context, sample_schema, n_samples, sop_data_generation_guidelines]
    text: |
      You are an expert AI assistant trained to generate high-quality synthetic datasets based on detailed Standard Operating Procedures (SOPs). Your task is to extract operational logic and decision conditions from the SOP to synthesize structured data that reflects the full range of possible cases.
      You will be provided with the following inputs:

      <sop_title>
      {{sop_title}}
      </sop_title>

      <sop_contents>
      {{sop_file_contents}}
      </sop_contents>

      {% if additional_context != "" %} 
      <additional_context>
      {{additional_context}}
      </additional_context>
      {% endif %}

      {% if sample_schema != "" %} 
      <schema>
      {{sample_schema}}
      </schema>
      {% endif %}

      Your task is to generate a synthetic dataset as follows:

      1. Interpretation: Carefully analyze the SOP to identify all decision points, validation steps, escalation conditions, and classifications. Use these to determine what fields must appear in the dataset and what values those fields may take.
      2. Schema-Adherence: Ensure the dataset reflects and expands upon the fields in the `<schema>` section, if provided. If no schema is provided, infer a realistic and complete schema from the SOP contents.
      3. Diversity of Cases:
        - Include both valid and invalid examples.
        - Cover edge cases, borderline values, and policy violation cases.
        - Ensure each step of the SOP's logic is reflected in at least some rows.
      4. Data Format:
        - Output the dataset within `<data></data>` tags only.
        - Use Python list-of-lists format including the header row, suitable for immediate use with CSV or Pandas.
        - Avoid any extra commentary, formatting, or explanations outside the `<data>` block.
      5. Validation Requirements:
        - All rows must be non-empty.
        - All columns must be complete and contain valid, well-formed values.
        - No unterminated strings, no newlines within strings.
        - Ensure consistency between field values (e.g., escalation should only be True when required conditions are met).
      6. Generate at least `{{n_samples}}` valid rows in total, spanning the full spectrum of SOP-defined conditions.
      7. Generate the dataset in one continuous pass without pausing, or asking for confirmation, or requesting interaction. Do not ask if you should continue.
      8. Do not include any explanatory notes, placeholder comments, ellipses, or messages like "[continuing...]" or "[...more rows below...]". Output only the full dataset rows within the `<data></data>` block without interruption. Any such messages will be treated as format violations.
      9. You must ensure that the data generated does not suffer from malformed node or string errors.
      10. Make sure there are no empty data columns or rows. Make sure there are NO unterminated string literals. Do not add new lines.
      11. For columns containing boolean, you must ensure that to use True or False so that it can be parsed using ast
      12. You must ensure you close any square brackets.
      
      {% if sop_data_generation_guidelines != "" %} 
      Additional requirements include:
      {{sop_data_generation_guidelines}}
      {% endif %}
      
      <data>
      [["field_1", "field_2", ..., "field_N"],
      ["value_1_1", "value_1_2", ..., "value_1_N"],
      ...
      ["value_k_1", "value_k_2", ..., "value_k_N"]]
      </data>
\end{lstlisting}

\subsection{SOP API Generation Prompt}
\label{appendix:sec:sop_api_generation}
\begin{lstlisting}[language=yaml, caption={Prompt Template YAML for converting user-provided task title, task context and sample data to APIs and Toolspec JSON}, breaklines=true, breakindent=0pt, columns=flexible, label={lst:api_prompt}]
prompts:
    - name: sop_api_generator
    parameters: [sop_file_contents, additional_context, sample_data]
    text: |
      You are an expert AI assistant specializing in API documentation and generation. Your task is to create sample API calls based on a Standard Operating Procedure (SOP) document. The goal is to produce accurate and comprehensive API documentation that captures all necessary steps and dependencies in the SOP.

      First, carefully read through the following SOP contents:

      <sop_contents>
      {{sop_file_contents}}
      </sop_contents>

      {% if additional_context != "" %} 
      Here is additional context you must adhere to for the SOP generation process: 
        <additional_context>
        {{additional_context}}
        </additional_context>
      {% endif %}

      {% if sample_data != "" %} 
      Here is a sample data for the SOP: 
        <sample_data>
        {{sample_data}}
        </sample_data>
      {% endif %}
      Now, conduct a thorough analysis of the SOP and additional context, focusing on the 'Main Procedure' section, and its sub-sections as well as the workflow. Wrap your analysis in <sop_breakdown> tags:

      <sop_breakdown>
      1. List all distinct tasks or operations in the 'Main Procedure' section.
      2. For each task, identify any external dependencies required for execution.
      3. Consider and note potential edge cases for each task.
      4. From the SOP, identify and list the key API parameters for each task.
      5. Plan your approach for API generation based on this analysis.
      6. Identify potential sources for input parameters (either from the SOP Input or from other API responses).
      7. Consider how different APIs might depend on each other and in what order they should be executed.
      8. For each potential API, list possible error scenarios and how they might be handled.
      9. from the sample data, list all the available columns.
      </sop_breakdown>
      
      <requirements>
      After completing your analysis, generate a set of tools/APIs based on the SOP. Follow these guidelines:
        1. Create a separate API for each distinct task or operation in the procedure.
        2. Ensure each tool or API is as granular or atomic as possible.
        3. Do not repeat tools or APIs for the same task.
        4. Capture all possible external dependencies the SOP needs to execute successfully.
        5. Ensure that input request parameters for each API are derived either from the SOP Input or from the response parameters (output) of other API calls.
        6. Make sure the APIs are complete and fully functional once implemented.
      
      For each API you generate, provide the following details:
        1. Name of the API
        2. API description
        3. Endpoint URI
        4. HTTP Method
        5. Request Body (including all necessary parameters) that is present in sample data within <sample_data> tags.
        6. Response (look for corresponding status columns for each API) that is present in sample data within <sample_data> tags.
        7  Do not add any parameters in Request and Response Body that is not present in sample data within <sample_data> tags.
        8. Dependencies (if this API depends on the output of other APIs, list them here)
        9. Potential error scenarios and suggested error handling
      </requirements>

      Wrap each tool/API inside <API></API> tags, and enclose the entire set of tools/APIs within <TOOLS></TOOLS> tags.

      Here is an example of the expected output structure:

      <TOOLS>
      <API>
      Name: [API Name]
      Description: [Brief description of the API function]
      Endpoint: [URI]
      Method: [HTTP Method]
      Request Body:
      {
        [JSON structure of request parameters]
      }
      Response:
      {
        [JSON structure of response data]
      }
      Dependencies: [List of dependent APIs, if any]
      Error Scenarios:
      - [Potential error 1]: [Suggested handling]
      - [Potential error 2]: [Suggested handling]
      </API>
      </TOOLS>

      It is OK for the <sop_breakdown> and <TOOLS> sections to be quite long, as they need to thoroughly cover all aspects of the SOP and resulting APIs.

  - name: generate_json_schema_for_each_api
    parameters: [sop_file_contents, additional_context, each_api_metadata, sample_data]
    text: |
      You are an expert AI assistant tasked with generating a JSON Schema for an API tool based on provided metadata. This tool is one of the many tools that will be used to execute a Standard Operating Procedure (SOP). Your goal is to create an accurate and detailed schema for this tool that aligns with the SOP and meets the specified requirements.

      First, please review the contents of the SOP:

      <sop_contents>
      {{sop_file_contents}}
      </sop_contents>

      {% if additional_context != "" %} 
      Here is additional context you must adhere to for the SOP generation process: 
        <additional_context>
        {{additional_context}}
        </additional_context>
      {% endif %}

      {% if sample_data != "" %} 
      Here is a sample data for the SOP: 
        <sample_data>
        {{sample_data}}
        </sample_data>
      {% endif %}

      Now, you will be provided with the metadata for a single API tool. Your task is to generate a JSON Schema for this tool. Here's how you should proceed:

      1. Carefully read and analyze the API metadata.
      2. Generate the following for the API tool:
        a. name: A concise, descriptive name for the tool.
        b. description: A clear explanation of the tool's purpose and functionality.
        c. inputSchema: A JSON schema format specifying the tool's 'input', including 'type', 'properties', and 'required' fields, all under the key 'json'.
      3. Ensure that the schema aligns with the overall SOP direction and any specified restrictions.

      Here is the API metadata you will be working with:

      <api_metadata>
      {{each_api_metadata}}
      </api_metadata>

      Before generating the final JSON Schema, wrap your analysis inside <sop_tool_analysis> tags:
      1. Quote relevant parts of the SOP that relate to the tool purpose.
      2. List potential input parameters based on the SOP and metadata.
      3. Consider any constraints or special considerations mentioned in the SOP.
      4. Outline how the tool aligns with the overall SOP direction.
      5. Identify potential edge cases or limitations of the tool.
      6. Describe how this tool fits into the larger workflow outlined in the SOP.

      After your analysis, generate the complete JSON Schema for the API tool. The schema should be formatted as an object with a "toolSpec" property, containing "name", "description", and "inputSchema" as shown in the example below:

      {
          "toolSpec": {
              "name": "example_tool_name",
              "description": "A clear description of the tool's purpose and functionality.",
              "inputSchema": {
                  "json": {
                      "type": "object",
                      "properties": {
                          "parameter_name": {
                              "type": "string",
                              "description": "A detailed description of this parameter."
                          }
                      },
                      "required": ["parameter_name"]
                  }
              }
          }
      }

      Remember to:
      - Ensure all required fields are listed in the "required" array within the input schema.
      - Provide detailed descriptions for each property in the input schema.
      - Use appropriate data types (string, integer, etc.) for each property.
      - Maintain consistency in formatting and structure.

      Present your final JSON Schema inside <SCHEMA></SCHEMA> tags.
\end{lstlisting}

\subsection{SOP API Function Code Prompt}
\label{appendix:sec:sop_api_code_generation}
\begin{lstlisting}[language=yaml, caption={Prompt Template YAML for implementing the functional code for an API, given sample data and a one-shot demonstration of an API along with its implementation and sample data as input}, breaklines=true, breakindent=0pt, columns=flexible, label={lst:apicode_prompt}]
prompts:
    - name: llm_coder
    parameters: [api, sample_data, api_example, coding_example, data_example]
    text: |
      You are an expert AI coding system who is capable of writing code in python using popular libraries.
      You are given the metadata of an API and a sample dataset. 
      Your task is to write a Python function that simulates the APIs.

      Here is an example of an API, dataset and its corresponding implementation.
      <example>
      Here is the api metedata:
      {{api_example}}

      Here is the dataset head in tsv:
      {{data_example}}

      Here is the corresponding code:
      {{coding_example}}
      <\example>

      <requirements>
        Requirement 1: Generate relevant Python code within <code></code> tags after closely studying the example provided in <example> tags.
        Requirement 2: You must write test cases as demonstrated in the example.
        Requirement 3: Import all required Python libraries. Use descriptive and meaningful variable names.
        Requirement 4: Name the function exactly as specified in the API specifications.
        Requirement 5: Conform strictly to the API definitions - the fields in the request must map to the function arguments, and the fields in the response must be returned.
        Requirement 6: Use only the existing columns present in the dataset.
        Requirement 7: Load the CSV dataset into memory before accessing the required fields.
        Requirement 8: Implement the process_tool_call() function exactly as demonstrated in the <example> tags.
        Requirement 9: Generate the code in one continuous pass without pausing, asking for confirmation, or requesting interaction. Do not ask if you should continue.
      </requirements>
      
      Following the above example, write code for the following API and dataset to complete the task successfully.
      <api>
      API metadata:
      {{api}}
      </api>   

      <sample_data>
      Here is the pandas DataFrame head (5 rows):
      {{sample_data}}
      </sample_data>

\end{lstlisting}

\section{Agent Prompt Templates}
\label{appendix:sec:agent_prompt_template}
\subsection{Function Calling Agent (FC-Agent)}
\label{appendix:sec:fc_agent_prompt_template}
\begin{lstlisting}[language=yaml, caption={Prompt Template YAML for FC-Agent to process input request following an SOP}, breaklines=true, breakindent=0pt, columns=flexible, label={lst:fc_agent_prompt}]
prompts:
  - name: simple_sop_agent_v1
    parameters: [SOP, USER_REQUEST]
    text: |
      You are an AI assistant responsible for processing user requests according to a specific Standard Operating Procedure (SOP). Your primary goal is to follow the SOP meticulously, ensuring consistency and accuracy in your responses.

      First, carefully read and internalize the following SOP:

      <sop>
      {{SOP}}
      </sop>

      Now, you will receive a user request to process. Your task is to handle this request by strictly adhering to the SOP provided above. Follow these steps:

      1. Read the user request carefully:
      <user_request>
      {{USER_REQUEST}}
      </user_request>

      2. Before responding, review the SOP once more to ensure you understand all steps. In <sop_analysis> tags:
        a. List the key points of the SOP.
        b. Identify any potential challenges or edge cases in applying the SOP to the user request.
        c. Consider possible interpretations of any ambiguous parts of the request.

      3. **CRITICAL:** Process the user request by following each step of the SOP in order. DO NOT skip any steps or add extra steps that are not in the SOP.

      4. For each step, perform the following actions:
        a. Log your progress in a <step_progress> tag, including the step number, description, and outcome (completed, failed, or pending).
        b. For critical steps, wrap your analysis inside <analysis> tags to show your calculations, validations, and reasoning.
        c. If a critical step cannot be completed, state the issue clearly and halt the process before proceeding to the next step.

      5. After every 3 steps, provide a checkpoint summary in <analysis> tags, highlighting critical validations performed and any discrepancies or issues encountered.

      6. If at any point you're unsure about how to proceed or if the user request doesn't align with the SOP, state this clearly in your response and halt the process.

      7. Once all steps are completed successfully, provide a final summary of the processed request in <final_response> tags, including all the key information as per the SOP.

      Example output structure:

      <step_progress>
      Step 1: [Description] - Outcome: [Completed/Failed/Pending]
      </step_progress>

      <analysis>
      [Detailed calculations, validations, or reasoning for critical steps]
      [Checkpoint summaries after every 3 steps]
      </analysis>

      <final_response>
      [Final response based on SOP processing]
      </final_response>

      Remember to adhere strictly to the SOP and provide clear, detailed explanations for each step of the process. It's OK for the <sop_analysis> section to be quite long.
\end{lstlisting}

\subsection{Prompt Template for React Agent}
\label{appendix:sec:react_agent_prompt_template}
\begin{lstlisting}[language=yaml, caption={System Prompt Template for React-Agent to process input request following an SOP}, breaklines=true, breakindent=0pt, columns=flexible, label={lst:react_agent_system_prompt}]
<system_prompt>You are a helpful assistant that follows Standard Operating Procedures (SOPs) to solve problems.

You will be given an SOP in an SOP block <sop> and a list of tools in a <tools> block.

Here are the tools:
\n\n
<tools>
{tools}
</tools>
    
<tool_input_guideline>
Follow these guidelines when using tools:
1. Examine the tool descriptions carefully to understand what each tool does
2. For each tool, provide all parameters in the correct order and format
3. Format tool inputs as valid JSON objects with all required fields
4. For tools that require nested objects, format them properly as JSON objects
5. Always check that your input includes all required parameters before using a tool
</tool_input_guideline>

Use the following format:
<format>
Question: the input question you must answer using the SOP and the current state of execution\n
Thought: the thought process that goes into answering the question\n
Action: the action to take, should be one of [{tool_names}]\n
Action Input: the input to the action as a valid JSON object with all required parameters\n
Observation: the result of the action\n
</format>

... (this Thought/Action/Action Input/Result can repeat N times)\n
eventually, once there are no more steps to execute or if the answer cannot be found based on the provided sop, tools and input.

Thought: I now know the final answer\n
Final Answer: the final answer to the original input question. Please make sure the final_decision is in the form:\n\n
<final_decision>FINAL_DECISION</final_decision>
<final_decision_reason>reason for final decision</final_decision_reason>
</system_prompt>

Begin:
\n\n
{input}

{agent_scratchpad}
\end{lstlisting}
\begin{lstlisting}[language=yaml, caption={A sample User Prompt Template for React-Agent to process input request following an SOP}, breaklines=true, breakindent=0pt, columns=flexible, label={lst:react_agent_user_prompt}]
<sop>\n\n{sop_text}\n\n</sop>\n\nHere is a video to process according to the SOP above:\n\n{formatted_input}\n\nPlease follow the SOP step by step to process this video.
\end{lstlisting}

%% file: kdd_tex_files/appendix_sop_example.tex
\newpage
\section{Examples of Real SOPs}
\label{appendix:real_sop}
\begin{figure*}[htb!]
 \centering 
  \includegraphics[width=\textwidth]{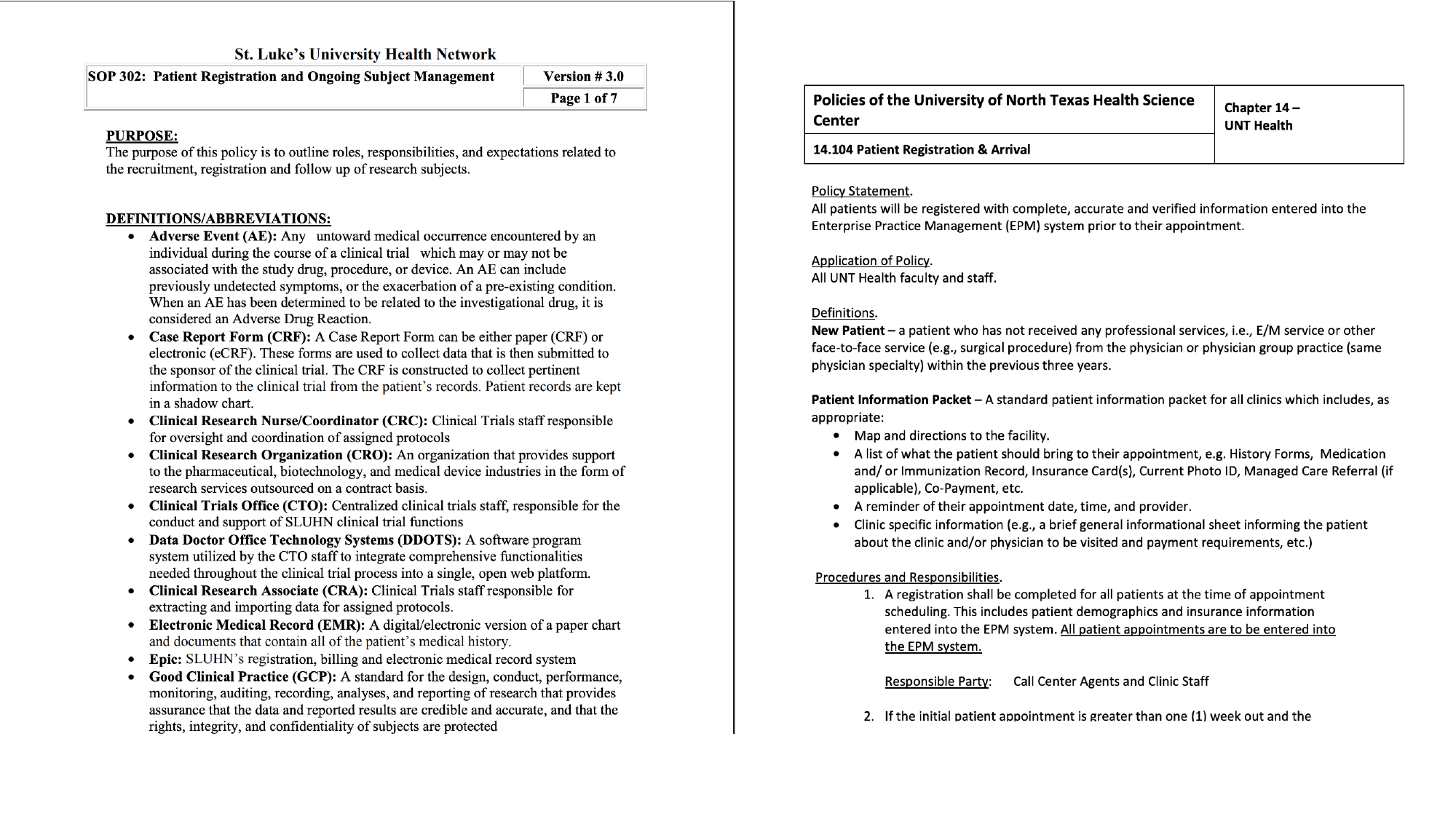}
    \caption{Examples of patient registration SOPs}
     \Description{Multi-page document excerpt showing several examples of patient registration standard operating procedures, including structured steps, headings, and form-like instructions.}
    \label{fig:example_sops}
\vspace{-0.25cm}
\end{figure*}

\section{SOP-Bench: Patient Intake and Registration}

This appendix provides an example of a Standard Operating Procedure (SOP) package for patient registration in a healthcare setting. It includes the main SOP document, JSON tool specifications, and sample patient data, illustrating how operational procedures are formalized and implemented in healthcare information systems.

\subsection{SOP: Patient Intake and Registration}

\begin{tcolorbox}[
  colback=gray!10,
  colframe=gray!50,
  width=0.5\textwidth,
  arc=2mm,
  title={\textbf{Patient Intake and Registration SOP - Part 1}},
  fonttitle=\bfseries,
  left=5pt,
  right=5pt
]
\footnotesize\begin{minipage}{\textwidth}
\begin{verbatim}
1. Purpose
This Standard Operating Procedure establishes a framework for intake 
and registration of new patients, encompassing demographic data 
collection, insurance verification, risk stratification, and pharmacy 
network validation in compliance with HIPAA regulations.

2. Scope
This procedure applies to all healthcare personnel involved in patient 
registration, including Patient Access Representatives, Insurance 
Verification Specialists, and Clinical Intake Coordinators.

3. Definitions
3.1 Protected Health Information (PHI): Individually identifiable 
    health information as defined by HIPAA
3.2 Benefits Coordination Protocol (BCP): Systematic verification of 
    primary and secondary insurance benefits
3.3 Clinical Risk Index (CRI): Composite score derived from medical 
    history and lifestyle factors
3.4 Pharmacy Benefit Management (PBM) Validation: Real-time 
    verification of prescription coverage
3.5 Network Adequacy Verification (NAV): Assessment of provider 
    network status
3.6 Risk Stratification Algorithm (RSA): Mathematical model for 
    patient risk assessment

4. Input
4.1 Required Documentation (some are optional)
- Government-issued photo identification (optional)
- Current insurance card(s)
- Complete medical history questionnaire
- Pharmacy benefit card
- Prior authorization documentation (optional)
- Release of medical records forms (optional)

4.2 System Requirements
- Electronic Health Record (EHR) access
- Insurance eligibility verification portal
- PBM interface connectivity
- Risk assessment calculation tool

5. Main Procedure

5.1 Insurance Validation Protocol
5.1.1 Primary Insurance Verification
- Access payer portal using provider credentials
- Verify coverage status and effective dates
- Document benefit levels and copayment requirements
- Confirm network participation status
- Record authorization requirements

5.1.2 Prescription Benefit Validation
- Query PBM database for current coverage
- Verify formulary compliance parameters
- Document prior authorization requirements
- Confirm specialty pharmacy protocols
- Record copayment structure

5.2 Risk Assessment Protocol
5.2.1 Lifestyle Risk Evaluation
\end{verbatim}
\end{minipage}\normalsize
\end{tcolorbox}

\begin{tcolorbox}[
  colback=gray!10,
  colframe=gray!50,
  width=0.52\textwidth,
  arc=2mm,
  title={\textbf{Patient Intake and Registration SOP - Part 2}},
  fonttitle=\bfseries,
  left=5pt,
  right=5pt
]
\footnotesize\begin{minipage}{\textwidth}
\begin{verbatim}

- Calculate smoking risk index (Never=0, Former=1, Current=2)
- Assess alcohol consumption (None=0, Occasional=1, Moderate=2, Heavy=3)
- Evaluate exercise patterns (5+ times=-1, 3-4 times=0, 1-2 times=1, None=2)
- Compute aggregate lifestyle score

5.2.2 Clinical Risk Assessment
- Review surgical history chronology
- Evaluate chronic condition severity
- Calculate comorbidity interaction score
- Generate weighted risk factor index

5.3 Patient Registration Protocol
5.3.1 Eligibility Confirmation
- Verify insurance_validation status = "valid"
- Confirm prescription_insurance_validation = "valid"
- Validate life_style_risk_level within parameters
- Assess overall_risk_level compatibility

5.3.2 Pharmacy Network Verification
- Query preferred pharmacy database
- Verify pharmacy_check status = "yes"
- Confirm pharmacy network participation
- Document special handling requirements

6. Output
6.1 Required Documentation
The procedure generates a verification report (json) with all results.
These include insurance validation check, prescription validation check, 
pharmacy check, life style risk, overall risk, and registration status.
All outputs must be archived in compliance with regulatory requirements.

6.2 System Updates
- Active patient status in EHR
- Documented risk levels
- Verified insurance records
- Confirmed pharmacy selection
- Completed registration status
\end{verbatim}
\end{minipage}\normalsize
\end{tcolorbox}

\newpage
\subsection{Sample Patient Data (Synthetically Generated)}

\begin{table}[!htbp]
\caption{Sample Patient Registration Data (First Five Records)}
\centering
\small
\resizebox{\textwidth}{!}{
\begin{tabular}{|p{1.6cm}|p{1.2cm}|p{1.2cm}|p{1.2cm}|p{0.8cm}|p{1.2cm}|p{1.1cm}|p{1.1cm}|p{1.1cm}|p{1.2cm}|p{1.2cm}|p{1.2cm}|}
\hline
\textbf{Patient ID} & \textbf{First Name} & \textbf{Last Name} & \textbf{DOB} & \textbf{Age} & \textbf{Insurance} & \textbf{Smoking} & \textbf{Alcohol} & \textbf{Exercise} & \textbf{Insurance Valid} & \textbf{Risk Level} & \textbf{Reg} \\
\hline
P100001 & John1 & Dove & 1980-05-15 & 43 & Blue Cross & Never & Occasional & 3-4 times & valid & low & success \\
\hline
P100002 & John2 & Dove & 1995-08-22 & 28 & Aetna & Current & Moderate & 1-2 times & valid & medium & success \\
\hline
P100003 & John3 & Dove & 1972-03-10 & 51 & United & Former & Heavy & None & valid & high & failure \\
\hline
P100004 & John4 & Dove & 1990-11-30 & 33 & Cigna & Never & None & 5+ times & valid & low & success \\
\hline
P100005 & John5 & Dove & 1965-07-25 & 58 & Medicare & Former & Occasional & 1-2 times & invalid & high & failure \\
\hline
\end{tabular}
}
\end{table}

\newpage
\subsection{Tool Specifications}

\begin{lstlisting}[language=json, caption={Tool Specifications in JSON Format}, breaklines=true, breakindent=0pt, columns=flexible, basicstyle=\small\ttfamily]

[
    {
        "toolSpec": {
            "name": "calculateLifestyleRisk",
            "description": "Calculates a patient's lifestyle risk level based on smoking status, alcohol consumption, and exercise frequency, following the standardized risk assessment protocol defined in the SOP.",
            "inputSchema": {
                "json": {
                    "type": "object",
                    "properties": {
                        "patient_id": {
                            "type": "string",
                            "description": "Unique identifier for the patient",
                            "pattern": "^P[0-9]{9}\$",
                            "example": "P123456789"
                        },
                        "smoking_status": {
                            "type": "string",
                            "enum": [
                                "Never",
                                "Former",
                                "Current"
                            ],
                            "description": "Patient's current smoking status, used to calculate smoking risk index"
                        },
                        "alcohol_consumption": {
                            "type": "string",
                            "enum": [
                                "None",
                                "Occasional",
                                "Moderate",
                                "Heavy"
                            ],
                            "description": "Patient's level of alcohol consumption frequency"
                        },
                        "exercise_frequency": {
                            "type": "string",
                            "enum": [
                                "None",
                                "1-2 times",
                                "3-4 times",
                                "5+ times"
                            ],
                            "description": "Weekly frequency of patient's exercise activities"
                        }
                    },
                    "required": [
                        "patient_id",
                        "smoking_status",
                        "alcohol_consumption",
                        "exercise_frequency"
                    ],
                    "additionalProperties": false
                }
            }
        }
    },
    {
        "toolSpec": {
            "name": "verifyPharmacy",
            "description": "Validates pharmacy network participation and verifies pharmacy details as part of the patient registration process",
            "inputSchema": {
                "json": {
                    "type": "object",
                    "properties": {
                        "patient_id": {
                            "type": "string",
                            "description": "Unique identifier for the patient",
                            "pattern": "^P[0-9]{9}\$",
                            "example": "P123456789"
                        },
                        "preferred_pharmacy_name": {
                            "type": "string",
                            "description": "Full legal name of the preferred pharmacy",
                            "minLength": 1,
                            "maxLength": 100,
                            "example": "CVS Pharmacy"
                        },
                        "preferred_pharmacy_address": {
                            "type": "string",
                            "description": "Complete street address of the pharmacy including city, state, and zip code",
                            "minLength": 5,
                            "maxLength": 200,
                            "example": "123 Main St, Boston, MA 02108"
                        },
                        "pharmacy_phone": {
                            "type": "string",
                            "description": "Contact phone number for the pharmacy in XXX-XXX-XXXX format",
                            "pattern": "^[0-9]{3}-[0-9]{3}-[0-9]{4}\$",
                            "example": "555-123-4567"
                        }
                    },
                    "required": [
                        "preferred_pharmacy_name",
                        "preferred_pharmacy_address",
                        "pharmacy_phone"
                    ],
                    "additionalProperties": false
                }
            }
        }
    }
]
\end{lstlisting}